\documentclass[11pt]{elsarticle}

\usepackage{framed,multirow}
\usepackage[utf8]{inputenc}
\usepackage[T1]{fontenc}
\usepackage{siunitx}
\usepackage{graphicx}
\usepackage{lineno}
\usepackage{hyperref}
\usepackage{subcaption}

\usepackage[ruled,vlined]{algorithm2e}
\usepackage{xcolor}

\SetKwInOut{KwRequire}{Require}
\SetKwInOut{KwEnsure}{Ensure}
\SetKwComment{Comment}{$\triangleright$\ }{}
\newcommand{\graycomment}[1]{\Comment{\textcolor{gray}{#1}}}

\newcounter{algline}

\DontPrintSemicolon

\usepackage{makecell}
\usepackage{subcaption}
\usepackage{hyperref,float}
\usepackage{booktabs}

\usepackage{tabu}
\usepackage{float}
\usepackage{placeins}
\usepackage{multirow} 
\usepackage{amssymb}
\usepackage{latexsym}
\usepackage{amsmath}
\usepackage[
left=3.8cm,
right=3.8cm,
top=4.8cm,
bottom=4.8cm
]{geometry}

\usepackage{caption}

\newcommand{\figslot}[4]{%
  \IfFileExists{#1}{\includegraphics[width=#2]{#1}}{%
    \fbox{\parbox[c][#3][c]{\dimexpr#2-2\fboxsep-2\fboxrule\relax}{\centering\small
      \textcolor{red}{\textbf{[Placeholder]}}\\[2pt]\texttt{\detokenize{#1}}\\[2pt]#4}}}}
\usepackage{amsthm}

\theoremstyle{plain}
\newtheorem{theorem}{Theorem}[section]
\newtheorem{proposition}[theorem]{Proposition}

\theoremstyle{definition}
\newtheorem{definition}[theorem]{Definition}

\theoremstyle{remark}

\usepackage{amsfonts}
\usepackage{float}
\usepackage{url}
\usepackage{xcolor}

\newcommand{\R}{\mathbb{R}}
\newcommand{\uu}{\mathbf{u}}

\newcommand{\SkewOp}{\operatorname{skew}}

\definecolor{newcolor}{rgb}{.8,.349,.1}

\journal{Computer Methods in Applied Mechanics and Engineering}

\usepackage[nolist]{acronym}
\begin{acronym}
\acro{PDE}[PDE]{partial differential equations}
\acro{BC}{boundary condition}
\acro{KANs}{Kolmogorov-Arnold networks}
\acro{KAN}{Kolmogorov-Arnold network}
\acro{SciML}{Scientific machine learning}
\acro{BCs}{\ac{BC}s}
\acro{MLPs}{multilayer perceptrons}
\acro{ML}{machine learning}
\acro{RBFs}{radial basis functions}
\acro{PINNs}{physics-informed neural networks}
\acro{DeepONet}{deep operator networks}
\acro{EDNNs}{evolutionary deep neural networks}
\acro{EvoKAN}{evolutionary \ac{KAN}}
\acro{GP}{Gaussian process}
\acro{SAV}{scalar auxiliary variable}
\acro{EDNN}{evolutionary deep neural networks}
\acro{EvoKAN}{evolutionary \ac{KAN}}
\acro{PINN}{physics-informed neural networks}
\acro{RTE}{radiative transfer equation}
\end{acronym}

\begin{document}
\begin{frontmatter}

\title{Machine Learning Moment Closure for the Boltzmann Transport Equation: Enforcing Hyperbolicity}

\author[1]{Bongseok Kim}

\author[2]{Jiahao Zhang}




\author[1,2]{Guang Lin\corref{cor1}}

\cortext[cor1]{Corresponding author.}
\ead{guanglin@purdue.edu}

\address[1]{School of Mechanical Engineering, Purdue University, West Lafayette, IN, USA}

\address[2]{Department of Mathematics, Purdue University, West Lafayette, IN, USA}

\begin{abstract}
The Boltzmann transport equation describes particle transport across a wide range of collisional regimes, but resolving its velocity dependence is computationally expensive.
Moment methods reduce this cost by evolving a finite number of velocity moments, at the price of a closure for the unresolved higher-order moments.
Data-driven closures can represent these moments accurately, yet without structural constraints they may produce complex characteristic speeds and, consequently, unstable simulations.
For the Boltzmann equation, learned closures with guaranteed hyperbolicity have so far been formulated in one spatial dimension, where the spectrum of a single matrix must be controlled; in multiple dimensions, hyperbolicity must hold for every direction simultaneously.
We develop, to the best of our knowledge, the first data-driven closure for multidimensional Hermite moment systems of the linear Boltzmann equation that is hyperbolic by construction.
The network learns a symmetric positive-definite symmetrizer that is shared by all spatial directions and restricted to the subspace that keeps the known lower-order transport equations exact; the unresolved rows then follow from a symmetric completion.
As a result, every directional Jacobian has real eigenvalues and a complete set of eigenvectors for every network output and at any truncation order.
The closure is trained only on spatial derivatives of unresolved moments from kinetic reference solutions, with no solver in the training loop, and is evolved by a partially conservative discontinuous Galerkin method.
In two-dimensional experiments in the transition and scattering-dominated regimes, a single learned third-order system with ten moments remains stable in all cases. 
\clearpage
Averaged over time, it is more accurate than the classical $P_7$ and $P_{15}$ truncations with $36$ and $136$ moments in every experiment, and its density error is about two orders of magnitude smaller than that of the $P_3$ truncation with the same moments.
\end{abstract}
\end{frontmatter}

\section{Introduction}

Kinetic equations describe a system between the particle and continuum levels, and they are used in rarefied gas dynamics, radiative transport, plasma physics, semiconductor transport, and other multiscale applications \cite{Alonso2015,Villani2002}. The Boltzmann equation is the central model of this class: its distribution function evolves in phase space and retains non-equilibrium information that macroscopic models discard \cite{DimarcoPareschi2014}. This information comes at a cost. A direct kinetic simulation must resolve the velocity variables in addition to physical space, which makes it far more expensive than solving a small system of macroscopic equations.

Moment methods reduce this cost by projecting the distribution onto a finite set of velocity basis functions. Grad's moment method established the Hermite expansion for gas kinetic theory \cite{Grad1949,Grad1958}, and moment hierarchies more generally provide a systematic route from kinetic to macroscopic descriptions \cite{Levermore1996}. Truncating the hierarchy, however, leaves the equation for the highest retained moment dependent on a moment that is not evolved. A closure relation is therefore required. Classical closures include polynomial and spherical-harmonic truncations, entropy-based models, and regularized moment systems \cite{Chandrasekhar1944,Alldredge2019}. They encode useful analytical structure, but their accuracy deteriorates away from the regime in which their assumptions hold.

A closure must also preserve the mathematical structure of the resulting first-order system. Hyperbolicity is essential: it guarantees real characteristic speeds and well-posed propagation of information. Grad-type systems are not globally hyperbolic; the equilibrium state of the classical 13-moment model, for instance, lies on the boundary of its hyperbolicity region \cite{CaiFanLi2014thirteen}. This observation motivated globally hyperbolic regularizations of moment systems \cite{CaiFanLi2013,CaiFanLi2014}. For data-driven closures, the issue is more severe. An unconstrained neural network can fit the training data well and still produce a Jacobian with complex eigenvalues at the states encountered during time integration.

Machine learning offers a flexible way to extract unresolved kinetic information from high-fidelity solutions. Han et al. trained generalized moments and their closure relations from kinetic data \cite{HanEtAl2019}. Neural closures have since been studied for Euler--Poisson and Vlasov-type systems \cite{BoisEtAl2022}, Landau fluid models \cite{MaEtAl2020}, and entropy-based kinetic closures \cite{PorteousEtAl2021,SchotthoferEtAl2025}. For radiative transfer, Huang et al. learned the spatial derivative of the unresolved moment rather than the moment itself \cite{HuangEtAl2022}, and subsequently enforced hyperbolicity with physical characteristic speeds \cite{HuangEtAl2023JSC} and global hyperbolicity \cite{HuangEtAl2023MMS}. Hyperbolic data-driven closures have also been developed for BGK moment systems \cite{ChristliebEtAl2024}, and data-driven closures for the linear Boltzmann equation have been studied in deterministic and stochastic settings \cite{HuangLiuQiWan2025}.

These studies show that accuracy alone is not a sufficient design criterion for a learned closure. A useful closure should satisfy two requirements. First, it should keep the known lower-order transport equations exact. Second, it should guarantee hyperbolicity for every network output, rather than penalize complex eigenvalues at sampled states. In multiple spatial dimensions, the second requirement is harder to meet. In one dimension, the closed system $\partial_t\uu+A(\uu)\partial_x\uu=0$ is hyperbolic if $A(\uu)$ is diagonalizable with real eigenvalues, and this is equivalent to the existence of a symmetrizer: if $A=R\Lambda R^{-1}$ with $\Lambda$ real, then $H=(RR^{\mathsf T})^{-1}\succ0$ satisfies $HA=(HA)^{\mathsf T}$. A one-dimensional closure therefore only has to control the spectrum of a single matrix. In two dimensions, the system $\partial_t\uu+A_x(\uu)\partial_x\uu+A_y(\uu)\partial_y\uu=0$ is hyperbolic only if the directional Jacobian
\begin{equation}
 A(\uu;\mathbf n)=n_xA_x(\uu)+n_yA_y(\uu)
 \label{eq:intro_directional_jacobian}
\end{equation}
is diagonalizable with real eigenvalues for every unit vector $\mathbf n=(n_x,n_y)$. This property does not follow from the coordinate matrices. For example,
\begin{equation}
 A_x=\begin{bmatrix}1&1\\0&-1\end{bmatrix},\qquad
 A_y=\begin{bmatrix}1&0\\-8&-1\end{bmatrix}
 \label{eq:intro_counterexample}
\end{equation}
both have the eigenvalues $\pm1$, whereas for $\mathbf n=(1,1)/\sqrt2$ the eigenvalues of $n_xA_x+n_yA_y$ satisfy $\lambda^2=-2$. Since $A_x$ and $A_y$ do not commute in general, the spectrum of $A(\uu;\mathbf n)$ is not determined by the spectra of $A_x$ and $A_y$, and a check at finitely many directions does not cover the others. A sufficient condition for all directions is a common symmetrizer $H(\uu)\succ0$ for which $HA_x$ and $HA_y$ are both symmetric, since then
\begin{equation}
 H^{1/2}A(\uu;\mathbf n)H^{-1/2}=H^{-1/2}\bigl(n_xHA_x+n_yHA_y\bigr)H^{-1/2}
 \label{eq:intro_common_symmetrizer}
\end{equation}
is symmetric for every $\mathbf n$. The symmetrizers obtained separately for $A_x$ and $A_y$ as in one dimension differ in general. Requiring a common one couples the closures of the two directions, and together with the known transport rows of both matrices, this coupling is the main difficulty addressed in this work. The learned closures with guaranteed hyperbolicity in \cite{HuangEtAl2023JSC,HuangEtAl2023MMS,ChristliebEtAl2024} are formulated in one spatial dimension and do not face this coupling. Data-driven approximations of entropy-based closures \cite{PorteousEtAl2021,SchotthoferEtAl2025} inherit hyperbolicity from a convex entropy, but they approximate the closure of a prescribed entropy rather than learning the unresolved moments from kinetic solutions. To the best of our knowledge, a data-driven closure that guarantees hyperbolicity of a multidimensional moment system of the Boltzmann equation for every network output, while keeping the known transport equations exact, has not been available.

In this work, we develop a neural moment closure for Hermite moment systems of the linear Boltzmann equation that meets both requirements by construction. The central object is a symmetric positive-definite inverse symmetrizer shared by both coordinate Jacobians. The known Hermite transport rows determine a linear subspace of admissible symmetrizers, and the network outputs coefficients in a fixed basis of this subspace. A normalized perturbation of the identity guarantees positive definiteness, and the unresolved rows are obtained by symmetric completion. Every directional Jacobian is then similar to a real symmetric matrix. The construction holds at any Hermite truncation order; only the dimensions and the subspace basis depend on the order.

The main contributions of this work are as follows:
\begin{enumerate}
    \item \textbf{Hyperbolicity in multiple dimensions by construction:} To the best of our knowledge, this is the first data-driven closure for multidimensional moment systems of the Boltzmann equation that guarantees hyperbolicity in every spatial direction for every network output. A common symmetrizer, restricted to the subspace compatible with the known Hermite rows, keeps the known transport equations exact, and the construction holds at any truncation order.
    \item \textbf{Stable DG simulation without solver-in-the-loop training:} Since the characteristic spectrum is real at every state, the learned system is coupled with a partially conservative discontinuous Galerkin (DG) method whose dissipation uses the exact characteristic speeds of the learned Jacobians. The closure is trained only on spatial derivatives of unresolved moments from kinetic reference data, and the resulting simulations remain stable in all experiments.
    \item \textbf{Low-order accuracy beyond high-order truncations:} With ten moments, the learned third-order system is, averaged over time, more accurate than the classical $P_7$ and $P_{15}$ truncations with $36$ and $136$ moments in all three experiments, and it reduces the density error of the $P_3$ truncation by about two orders of magnitude.
\end{enumerate}

The numerical workflow keeps kinetic reference generation separate from moment evolution. A discrete-velocity solution of the linear Boltzmann equation provides reference moments and their spatial derivatives, and the closure is trained to reproduce the derivatives of the unresolved moments. The moment system is evolved by a modal DG method that uses conservative fluxes in the known lower-order equations and path fluctuations in the highest-order equations. Transport is advanced by strong-stability-preserving Runge--Kutta time integration \cite{ShuOsher1988}, and collisions are updated exactly. The remainder of the paper is organized as follows. Section~\ref{sec:preliminaries} reviews the linear Boltzmann equation and its Hermite moment system. Section~\ref{sec:methodology} presents the hyperbolic neural closure, its training, and the DG discretization. Section~\ref{Sec:numerical_experiments} reports numerical experiments, and Section~\ref{sec:conclusion} concludes the paper.

\section{Preliminaries}
\label{sec:preliminaries}
\subsection{Boltzmann transport equation}

We consider the linear Boltzmann transport equation in two spatial and two velocity dimensions. Let $\mathbf{x}=(x,y)\in\Omega\subset\R^2$, $\mathbf{v}=(v_x,v_y)\in\R^2$, and let $f(t,\mathbf{x},\mathbf{v})$ denote the particle distribution. The governing equation is
\begin{equation}
    \partial_t f+v_x\partial_x f+v_y\partial_y f=Q(f).
    \label{eq:2d_bte}
\end{equation}
We use isotropic linear scattering toward a fixed normalized equilibrium,
\begin{align}
    Q(f)&=\sigma\bigl[M_2(\mathbf{v})\rho_f(t,\mathbf{x})-f(t,\mathbf{x},\mathbf{v})\bigr],
    \qquad \sigma>0,\label{eq:2d_linear_collision}\\
    \rho_f(t,\mathbf{x})&:=\int_{\R^2}f(t,\mathbf{x},\mathbf{v})\,d\mathbf{v},
    \qquad M_2(\mathbf{v}):=\frac1\pi e^{-(v_x^2+v_y^2)}.
\end{align}
Here $\sigma$ is a constant collision frequency and $\int_{\R^2}M_2\,d\mathbf{v}=1$. Consequently, $\int_{\R^2}Q(f)\,d\mathbf{v}=0$: scattering preserves density while relaxing the velocity distribution toward $M_2\rho_f$. Initial and boundary conditions are specified for each numerical experiment. The fixed equilibrium permits a fixed Hermite basis and isolates the transport closure from nonlinear collision modeling.

\subsection{Moment method}

Let $M_1(s)=\pi^{-1/2}e^{-s^2}$ and let $H_k$ be the normalized Hermite polynomial of degree $k$ with
\begin{equation}
    \int_{\R}H_k(s)H_\ell(s)M_1(s)\,ds=\delta_{k\ell},
    \qquad H_0=1,\quad H_1(s)=\sqrt2\,s.
\end{equation}
The tensor-product polynomials $\Psi_{ab}(\mathbf{v})=H_a(v_x)H_b(v_y)$ are orthonormal with respect to $M_2=M_1(v_x)M_1(v_y)$. Define the moments
\begin{equation}
    m_{ab}(t,\mathbf{x}):=\int_{\R^2}f(t,\mathbf{x},\mathbf{v})\Psi_{ab}(\mathbf{v})\,d\mathbf{v},
    \qquad a,b\in\mathbb N_0.
    \label{eq:2d_moment_definition}
\end{equation}
In particular, $m_{00}=\rho_f$. The Hermite recurrence reads
\begin{equation}
    sH_k(s)=\sqrt{\frac{k+1}{2}}H_{k+1}(s)+\sqrt{\frac{k}{2}}H_{k-1}(s),
    \label{eq:hermite_recurrence}
\end{equation}
where the second term is omitted for $k=0$. Multiplying Eq.~\eqref{eq:2d_bte} by $\Psi_{ab}$ and integrating over velocity gives
\begin{align}
    \partial_t m_{ab}
    &+\sqrt{\frac{a+1}{2}}\partial_xm_{a+1,b}
     +\sqrt{\frac a2}\partial_xm_{a-1,b}\nonumber\\
    &+\sqrt{\frac{b+1}{2}}\partial_ym_{a,b+1}
     +\sqrt{\frac b2}\partial_ym_{a,b-1}
     =\sigma(\delta_{a0}\delta_{b0}m_{00}-m_{ab}).
    \label{eq:2d_moment_eq}
\end{align}
Terms with negative indices are omitted. The source follows from $\int M_2\Psi_{ab}\,d\mathbf{v}=\delta_{a0}\delta_{b0}$. For example, the first three equations are
\begin{align}
    \partial_tm_{00}+\frac1{\sqrt2}(\partial_xm_{10}+\partial_ym_{01})&=0,\\
    \partial_tm_{10}+\frac1{\sqrt2}\partial_xm_{00}+\partial_xm_{20}
        +\frac1{\sqrt2}\partial_ym_{11}&=-\sigma m_{10},\\
    \partial_tm_{01}+\frac1{\sqrt2}\partial_xm_{11}
        +\frac1{\sqrt2}\partial_ym_{00}+\partial_ym_{02}&=-\sigma m_{01}.
\end{align}
Thus each layer of fixed total degree couples to the adjacent layers.

For a truncation order $N\ge1$, retain the index set
\begin{equation}
    \mathcal I_N:=\{(a,b)\in\mathbb N_0^2:a+b\le N\},
    \qquad M_N:=|\mathcal I_N|=\frac{(N+1)(N+2)}2.
\end{equation}
Order the moments by increasing total degree and, within degree $k$, as $(k,0),(k-1,1),\ldots,(0,k)$. Let $r_N(a,b)$ be the corresponding one-based index in $\mathcal I_N$. The resolved state and its partition are
\begin{equation}
    \uu_N=(m_{00},m_{10},m_{01},\ldots,m_{N0},\ldots,m_{0N})^{\mathsf T}
    =\begin{bmatrix}\uu_{<N}\\\uu_N^{\mathrm{high}}\end{bmatrix}\in\R^{M_N},
    \label{eq:general_state_partition}
\end{equation}
where $p_N:=\dim\uu_{<N}=N(N+1)/2$ and $q_N:=\dim\uu_N^{\mathrm{high}}=N+1$. All equations with $a+b<N$ involve only resolved moments. The $q_N$ equations with $a+b=N$ additionally involve the $N+2$ unresolved moments of total degree $N+1$.

To express this structure, define the truncated Hermite multiplication matrices $T_{x,N},T_{y,N}\in\R^{M_N\times M_N}$ by
\begin{align}
 (T_{x,N})_{r_N(a,b),r_N(c,d)}
 &=\sqrt{\frac{a+1}{2}}\delta_{c,a+1}\delta_{d,b}
   +\sqrt{\frac a2}\delta_{c,a-1}\delta_{d,b},\\
 (T_{y,N})_{r_N(a,b),r_N(c,d)}
 &=\sqrt{\frac{b+1}{2}}\delta_{c,a}\delta_{d,b+1}
   +\sqrt{\frac b2}\delta_{c,a}\delta_{d,b-1},
\end{align}
for $(a,b),(c,d)\in\mathcal I_N$. Both matrices are symmetric. If $\mathbf e_r$ denotes the $r$th basis vector in $\R^{M_N}$, the exact truncated hierarchy is
\begin{align}
 &\partial_t\uu_N+T_{x,N}\partial_x\uu_N+T_{y,N}\partial_y\uu_N\nonumber\\
 &\quad+\sum_{a+b=N}\mathbf e_{r_N(a,b)}
 \left(\sqrt{\frac{a+1}{2}}\partial_xm_{a+1,b}
       +\sqrt{\frac{b+1}{2}}\partial_ym_{a,b+1}\right)
 =S_\sigma\uu_N,\label{eq:2d_unclosed_system}\\
 S_\sigma&:=\operatorname{diag}(0,-\sigma,\ldots,-\sigma)\in\R^{M_N\times M_N}.
\end{align}
A classical truncation sets the unresolved moments to zero. A learned closure instead approximates their contribution to Eq.~\eqref{eq:2d_unclosed_system}.

For a differentiable scalar closure $m_{cd}=g_{cd}(\uu_N)$, $c+d=N+1$, the chain rule gives a quasilinear system
\begin{equation}
 \partial_t\uu_N+A_{x,N}(\uu_N)\partial_x\uu_N
                 +A_{y,N}(\uu_N)\partial_y\uu_N=S_\sigma\uu_N.
 \label{eq:2d_closed_system}
\end{equation}
For example,
\begin{equation}
 A_{x,N}=T_{x,N}+\sum_{a+b=N}\sqrt{\frac{a+1}{2}}
 \mathbf e_{r_N(a,b)}\bigl(\nabla_{\uu_N}g_{a+1,b}\bigr)^{\mathsf T},
 \label{eq:2d_scalar_closure_matrix}
\end{equation}
with the analogous expression for $A_{y,N}$. The proposed method parameterizes these transport matrices directly, without assuming that every learned row is the gradient of a scalar closure.

In either case, the first $p_N$ rows must remain equal to the known Hermite rows,
\begin{equation}
 F_i^{(N)}:=(T_{i,N})_{1:p_N,:}
 =\begin{bmatrix}F_{i,LL}^{(N)}&F_{i,LH}^{(N)}\end{bmatrix},
 \qquad i\in\{x,y\}.
 \label{eq:general_fixed_rows}
\end{equation}
Therefore the matrices to be closed have the structure
\begin{equation}
 A_{i,N}=\begin{bmatrix}F_{i,LL}^{(N)}&F_{i,LH}^{(N)}\\C_{i,N}&D_{i,N}\end{bmatrix},
 \quad C_{i,N}\in\R^{q_N\times p_N},\quad D_{i,N}\in\R^{q_N\times q_N},
 \label{eq:general_A_block}
\end{equation}
where the leading block is symmetric,
\begin{equation}
 F_{i,LL}^{(N)}=\left(F_{i,LL}^{(N)}\right)^{\mathsf T},\qquad i\in\{x,y\},
 \label{eq:general_FLL_symmetric}
\end{equation}
since $T_{x,N}$ and $T_{y,N}$ are symmetric. Only the last $q_N$ rows are closure-dependent.

\subsection{Hyperbolicity}
\label{sec:hyperbolicity}

Our primary goal is a closure for which the closed moment system remains hyperbolic for every network output. To state this requirement precisely, we consider a general first-order system in $d$ spatial dimensions,
\begin{equation}
 \frac{\partial\uu}{\partial t}+\sum_{i=1}^{d}A_i(\uu)\frac{\partial\uu}{\partial x_i}=S\uu,
 \qquad
 \mathbf x\in\Omega\subset\R^d,
 \qquad
 t\in(0,T),
 \label{eq:prelim_quasilinear}
\end{equation}
where $\uu=(u_1,\dots,u_p)^{\mathsf T}\in\mathcal D\subset\R^p$ is the state vector, $A_i(\uu)\in\R^{p\times p}$ are the transport matrices, and $S\in\R^{p\times p}$ is a relaxation matrix. For a system of conservation laws with fluxes $\mathbf f_i(\uu)$, the transport matrices are the flux Jacobians $A_i(\uu)=\nabla_{\uu}\mathbf f_i(\uu)$. For any unit vector $\mathbf n=(n_1,\dots,n_d)^{\mathsf T}\in\mathbb S^{d-1}$, we define the directional Jacobian $A(\uu;\mathbf n):=\sum_{i=1}^dn_iA_i(\uu)$.

\begin{definition}[Hyperbolicity]
\label{def:hyperbolicity}
The system \eqref{eq:prelim_quasilinear} is hyperbolic on $\mathcal D$ if, for every $\uu\in\mathcal D$ and every unit vector $\mathbf n\in\mathbb S^{d-1}$, the matrix $A(\uu;\mathbf n)$ has $p$ real eigenvalues and a complete set of linearly independent eigenvectors.
\end{definition}

The eigenvalues of $A(\uu;\mathbf n)$ are the characteristic speeds in the direction $\mathbf n$. The relaxation term is of lower order and does not enter Definition~\ref{def:hyperbolicity}. Hyperbolicity is a property of every direction simultaneously, so it is not implied by hyperbolicity of each coordinate matrix $A_i$ alone. A convenient sufficient condition is the existence of one symmetrizer shared by all coordinate directions.

\begin{proposition}[Common symmetrizer]
\label{prop:common_symmetrizer}
Suppose that, for every $\uu\in\mathcal D$, there exists a symmetric positive definite matrix $H(\uu)$, independent of $\mathbf n$, such that $H(\uu)A_i(\uu)$ is symmetric for every $i$. Then the system \eqref{eq:prelim_quasilinear} is hyperbolic on $\mathcal D$.
\end{proposition}

\begin{proof}
For every unit vector $\mathbf n$, the matrix $H A(\uu;\mathbf n)=\sum_{i=1}^dn_iHA_i$ is symmetric. Hence $H^{1/2}A(\uu;\mathbf n)H^{-1/2}=H^{-1/2}\bigl(HA(\uu;\mathbf n)\bigr)H^{-1/2}$ is real and symmetric, and $A(\uu;\mathbf n)$ is similar to it. Therefore $A(\uu;\mathbf n)$ has real eigenvalues and a complete set of eigenvectors.
\end{proof}

For the Hermite moment system \eqref{eq:2d_closed_system}, $d=2$, $p=M_N$, $A_i=A_{i,N}$, and $S=S_\sigma$. The classical $P_N$ truncation keeps $A_{i,N}=T_{i,N}$, which is symmetric, so $H=I_{M_N}$ is a common symmetrizer and the truncated system is hyperbolic. A learned closure modifies the last $q_N$ rows, and an unconstrained modification generally destroys this property. If each coordinate matrix is symmetrized by a different matrix, the directional Jacobian $A_N(\uu_N;\mathbf n)$ may still have complex eigenvalues for oblique directions. The construction in Section~\ref{sec:common_symmetrizer} therefore enforces the condition of Proposition~\ref{prop:common_symmetrizer} directly.

\section{Proposed Methodology: Hyperbolic neural moment closure for BTE}
\label{sec:methodology}
\subsection{Common symmetrizer and symmetric completion}
\label{sec:common_symmetrizer}

For a unit direction $\mathbf n=(n_x,n_y)\in\mathbb S^1$, define
\begin{equation}
 A_N(\uu_N;\mathbf n):=n_xA_{x,N}(\uu_N)+n_yA_{y,N}(\uu_N).
\end{equation}
By Definition~\ref{def:hyperbolicity}, the closed system \eqref{eq:2d_closed_system} is hyperbolic if this matrix has real eigenvalues and a complete eigenbasis for every direction. Following Proposition~\ref{prop:common_symmetrizer}, we construct one symmetric positive definite matrix $H_N(\uu_N)$ shared by both directions, with
\begin{equation}
 A_{i,N}=S_{i,N}H_N,\qquad S_{i,N}=S_{i,N}^{\mathsf T},\qquad H_N\succ0.
 \label{eq:2d_factorization}
\end{equation}
Then $H_NA_{i,N}=H_NS_{i,N}H_N$ is symmetric. We work with the inverse symmetrizer $W_N:=H_N^{-1}$, so $S_{i,N}=A_{i,N}W_N$.

To identify the admissible inverse symmetrizers, note that once $W_N$ is chosen, the first $p_N$ rows of $S_{i,N}=A_{i,N}W_N$ are fixed by Eq.~\eqref{eq:general_fixed_rows} and equal $F_i^{(N)}W_N$. Since $S_{i,N}$ must be symmetric, the leading $p_N\times p_N$ block of $F_i^{(N)}W_N$ must itself be symmetric. These are linear conditions on $W_N$, which define a compatible subspace; the remaining lower-right blocks of $S_{x,N}$ and $S_{y,N}$ can then be completed freely. With $\SkewOp(X):=\tfrac12(X-X^{\mathsf T})$, the following theorem makes this construction precise for every truncation order.

\begin{theorem}[Symmetric completion with a common symmetrizer]
\label{thm:order_generic_completion}
Let $N\ge1$ and let the fixed Hermite rows $F_i^{(N)}$ be defined by Eq.~\eqref{eq:general_fixed_rows}. Partition a symmetric matrix $W\in\mathbb S^{M_N}$ as
\begin{equation}
    W=
    \begin{bmatrix}
        U_N&V_N\\
        V_N^{\mathsf T}&Z_N
    \end{bmatrix},
    \label{eq:general_W_partition}
\end{equation}
with blocks conforming to the partition in Eq.~\eqref{eq:general_state_partition}, and define
\begin{align}
    X_{i,N}(W)
    &:=F_{i,LL}^{(N)}U_N+F_{i,LH}^{(N)}V_N^{\mathsf T},
    \label{eq:general_X}\\
    Y_{i,N}(W)
    &:=F_{i,LL}^{(N)}V_N+F_{i,LH}^{(N)}Z_N.
    \label{eq:general_Y}
\end{align}
Define the linear compatibility operator
$\mathcal C_N(W) :=\left( \SkewOp(X_{x,N}(W)), \SkewOp(X_{y,N}(W)) \right)$,
and the compatible subspace
$\mathcal L_N:=\ker\mathcal C_N\subset\mathbb S^{M_N}$.
Let $\{B_k^{(N)}\}_{k=1}^{d_N}$ be any basis of $\mathcal L_N$. Suppose a neural network produces coefficients $\theta_k^{(N)}(\uu_N)$, a scalar $\widehat\tau_N(\uu_N)$, and symmetric directional matrices
$R_{i,N}=R_{i,N}^{\mathsf T}\in\R^{q_N\times q_N}, \, \, i\in\{x,y\}$.
Set
\begin{align}
    \widetilde W_N(\uu_N)
    &:=\sum_{k=1}^{d_N}\theta_k^{(N)}(\uu_N)B_k^{(N)},\\
    E_N(\uu_N)
    &:=\frac{\widetilde W_N(\uu_N)}{1+\|\widetilde W_N(\uu_N)\|_F},\\
    \tau_N(\uu_N)
    &:=\operatorname{softplus}(\widehat\tau_N(\uu_N))+\varepsilon,
    \qquad \varepsilon>0,
\end{align}
and, for a fixed $0<\eta<1$,
\begin{equation}
    W_N(\uu_N) :=\tau_N(\uu_N)\left(I_{M_N}+\eta E_N(\uu_N)\right).
    \label{eq:2d_Wspd}
\end{equation}
Let $H_N=W_N^{-1}$ and define
\begin{equation}
    S_{i,N}:=
    \begin{bmatrix}
        X_{i,N}(W_N)&Y_{i,N}(W_N)\\
        Y_{i,N}(W_N)^{\mathsf T}&R_{i,N}
    \end{bmatrix},
    \qquad
    A_{i,N}:=S_{i,N}H_N.
    \label{eq:general_SA_theorem}
\end{equation}
Then the following properties hold for every state $\uu_N$:
\begin{enumerate}
    \item $W_N\in\mathcal L_N$ and $W_N\succ0$.
    \item the known Hermite transport rows are preserved exactly,
    $(A_{i,N})_{1:p_N,:}=F_i^{(N)}, \, \,  i\in\{x,y\}$.
    \item for every unit direction $\mathbf n=(n_x,n_y)^{\mathsf T}\in\mathbb S^1$, the directional Jacobian
    \[
    A_N(\uu_N;\mathbf n):=n_xA_{x,N}+n_yA_{y,N}
    \]
    has only real eigenvalues and a complete eigenbasis.
\end{enumerate}
Hence the construction defines a hyperbolic neural completion for every truncation order, without imposing additional fixed entries on the closure-dependent rows beyond the algebraic conditions required by simultaneous symmetrizability.
\end{theorem}

\begin{proof}
By Eq.~\eqref{eq:general_FLL_symmetric},
$X_{i,N}(I_{M_N})=F_{i,LL}^{(N)} =\left(F_{i,LL}^{(N)}\right)^{\mathsf T}, \, \, i\in\{x,y\}$,
so $\mathcal C_N(I_{M_N})=0$ and hence $I_{M_N}\in\mathcal L_N$. The matrix $\widetilde W_N$ is a linear combination of basis matrices of $\mathcal L_N$, so $\widetilde W_N\in\mathcal L_N$. Since the normalization factor $[1+\|\widetilde W_N\|_F]^{-1}$ is scalar, $E_N\in\mathcal L_N$ as well. Therefore $I_{M_N}+\eta E_N\in\mathcal L_N$, and multiplication by the scalar $\tau_N$ yields $W_N\in\mathcal L_N$.

The normalization gives
$\|E_N\|_F =\frac{\|\widetilde W_N\|_F}{1+\|\widetilde W_N\|_F}<1$,
so $\|E_N\|_2\le\|E_N\|_F<1$. Moreover, $E_N$ is symmetric because $\mathcal L_N\subset\mathbb S^{M_N}$. Thus $\lambda_{\min}(E_N)\ge-\|E_N\|_2$. For every nonzero $z\in\R^{M_N}$,
\begin{align}
    z^{\mathsf T}W_Nz
    &=\tau_N z^{\mathsf T}(I_{M_N}+\eta E_N)z\nonumber\\
    &=\tau_N\left(\|z\|_2^2+\eta z^{\mathsf T}E_Nz\right)\nonumber\\
    &\ge\tau_N\left(1+\eta\lambda_{\min}(E_N)\right)\|z\|_2^2\nonumber\\
    &\ge\tau_N\left(1-\eta\|E_N\|_2\right)\|z\|_2^2\nonumber\\
    &>\tau_N(1-\eta)\|z\|_2^2>0.
\end{align}
Hence $W_N\succ0$. Since $W_N$ is symmetric positive definite, its inverse exists and has strictly positive eigenvalues, so $H_N=W_N^{-1}\succ0$.

Because $W_N\in\mathcal L_N$, the compatibility conditions
$X_{x,N}=X_{x,N}^{\mathsf T}$ and $ X_{y,N}=X_{y,N}^{\mathsf T}$
hold exactly. Therefore each $S_{i,N}$ in Eq.~\eqref{eq:general_SA_theorem} is symmetric. Moreover,
\begin{equation}
    \begin{bmatrix}X_{i,N}&Y_{i,N}\end{bmatrix}
    =F_i^{(N)}W_N,
\end{equation}
and hence
$(A_{i,N})_{1:p_N,:} =F_i^{(N)}W_NH_N =F_i^{(N)}$,
which proves exact preservation of all known rows.

Finally, for any $\mathbf n\in\mathbb S^1$,
$A_N(\uu_N;\mathbf n) =S_N(\uu_N;\mathbf n)H_N$ with $ S_N(\uu_N;\mathbf n):=n_xS_{x,N}+n_yS_{y,N}$.
Since $S_{x,N}$ and $S_{y,N}$ are symmetric, $S_N(\uu_N;\mathbf n)$ is symmetric. Therefore
$H_N^{1/2}A_N(\uu_N;\mathbf n)H_N^{-1/2} =H_N^{1/2}S_N(\uu_N;\mathbf n)H_N^{1/2}$
is real and symmetric. Thus $A_N(\uu_N;\mathbf n)$ is similar to a real symmetric matrix and has real eigenvalues with a complete eigenbasis.
\end{proof}

This result controls the reality and completeness of the characteristic spectrum. It does not, by itself, impose an upper bound on characteristic speeds or guarantee stability of a particular discretization.

Since the time step of an explicit discretization is limited by the largest characteristic speed, we further restrict the symmetric completion so that these speeds remain bounded, without changing the known rows or adding a spectral penalty to the training loss. Write $W_N=LL^{\mathsf T}$ with $L$ lower triangular and define
\begin{equation}
 B_i:=L^{-1}S_{i,N}L^{-\mathsf T}
 =\begin{bmatrix}E_i&G_i\\G_i^{\mathsf T}&J_i\end{bmatrix},
 \qquad
 [E_i\ G_i]=L_{LL}^{-1}F_i^{(N)}W_NL^{-\mathsf T},
 \qquad i\in\{x,y\},
 \label{eq:bounded_similarity}
\end{equation}
where $L_{LL}$ is the leading $p_N\times p_N$ block of $L$. The matrix $B_i$ is symmetric and similar to $A_{i,N}$, and only $J_i\in\mathbb S^{q_N}$ remains to be chosen. Given $c_0>0$, $\delta_c>0$, and $\widehat K_i\in\mathbb S^{q_N}$, let $c=c_i(\uu_N):=\max\{c_0,(1+\delta_c)\|[E_i\ G_i]\|_2\}$ and
\begin{align}
 V_i&:=c\left[I_{q_N}-G_i^{\mathsf T}(c^2I_{p_N}-E_i^2)^{-1}G_i\right]=Z_iZ_i^{\mathsf T},\nonumber\\
 J_i&:=-G_i^{\mathsf T}E_i(c^2I_{p_N}-E_i^2)^{-1}G_i+Z_iK_iZ_i^{\mathsf T},\qquad
 K_i:=\frac{\widehat K_i}{1+\|\widehat K_i\|_F},
 \label{eq:speed_bounded_completion}
\end{align}
with $Z_i$ the Cholesky factor. Since $c>\|[E_i\ G_i]\|_2$, $V_i\succ0$, and the Schur complements of $cI_{M_N}\mp B_i$ are $V_i\mp Z_iK_iZ_i^{\mathsf T}\succ0$ because $\|K_i\|_2<1$. Hence every eigenvalue of $A_{i,N}$ lies in $(-c_i,c_i)$, and the lower-right block of $S_{i,N}=LB_iL^{\mathsf T}$ defines $R_{i,N}$ in Eq.~\eqref{eq:general_SA_theorem}, so Theorem~\ref{thm:order_generic_completion} applies for every $\widehat K_i$. We use $c_0=3.8897$, the largest characteristic speed of the $P_{11}$ truncation, and $\delta_c=0.02$.

\subsection{Neural parameterization and a priori training}
\label{sec:neural_parameterization}

We now parameterize the admissible inverse symmetrizer and directional completions, and train the resulting matrices using unresolved spatial derivatives from kinetic reference data.

For each order $N$, the basis $\{B_k^{(N)}\}_{k=1}^{d_N}$ of $\mathcal L_N$ in Theorem~\ref{thm:order_generic_completion} is computed once offline. We construct $F_x^{(N)}$ and $F_y^{(N)}$ from the Hermite recurrence, represent $\mathcal C_N$ on the independent entries of a symmetric matrix, and obtain a basis of its kernel by an SVD or a rank-revealing QR factorization, where $d_N=\dim\mathcal L_N$. No nullspace factorization is needed during training or moment evolution.

\begin{figure}[htp!]
\centering
\begin{subfigure}[b]{0.95\linewidth}
    \centering
    \includegraphics[width=\linewidth]{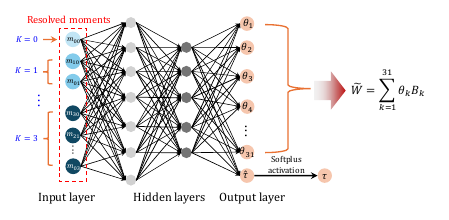}
    \subcaption{Shared symmetrizer network.}
    \label{fig:hn_architecture_symmetrizer}
\end{subfigure}
\hfill
\begin{subfigure}[b]{0.95\linewidth}
    \centering
    \includegraphics[width=\linewidth]{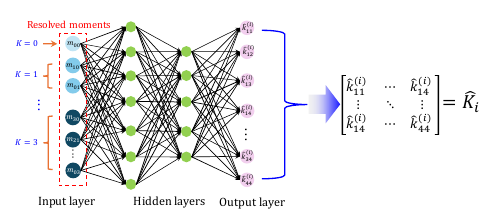}
    \subcaption{Directional networks.}
    \label{fig:hn_architecture_directional}
\end{subfigure}
\caption{Structure of the HN closure for order $N$. Standardized resolved moments enter (a) a shared symmetrizer network, which outputs the coefficients $\theta_k^{(N)}$ in the basis of $\mathcal L_N$ and the scale $\widehat\tau_N$, and (b) two directional networks, which output the entries $\widehat k^{(i)}_{jl}$, $j\le l$, of the symmetric matrices $\widehat K_i$, $i\in\{x,y\}$. The inverse symmetrizer $W_N$ is formed by Eq.~\eqref{eq:2d_Wspd}, the directional completions by Eqs.~\eqref{eq:bounded_similarity}--\eqref{eq:speed_bounded_completion}, and the transport matrices by $A_{i,N}=S_{i,N}H_N$. The known Hermite rows are never modified. During training, the learned rows are contracted with reference derivatives to predict the unresolved derivatives in Eq.~\eqref{eq:unresolved_gradient_predictions}; during simulation, the matrices enter the DG discretization of Section~\ref{sec:jacobian_first_numerics}.}
\label{fig:hn_architecture}
\end{figure}

Regarding the network outputs, three networks, with collective parameters $\phi$, take standardized resolved moment values as inputs. For component $j$, the network input is $(u_j-\mu_j)/s_j^{\mathrm{in}}$, where $\mu_j$ and $s_j^{\mathrm{in}}$ are the mean and standard deviation over the training set. A shared network outputs the $d_N$ coefficients $\theta_k^{(N)}$ and the scalar $\widehat\tau_N$ in Theorem~\ref{thm:order_generic_completion}, while separate directional networks output the independent entries of symmetric $q_N\times q_N$ matrices $\widehat K_x,\widehat K_y$, which determine the directional completions. The combined output dimension is $d_N+1+q_N(q_N+1)$. Here $\operatorname{softplus}(s)=\log(1+e^s)$ and $\eta$ is fixed. Besides positive definiteness, the bound $\|E_N\|_2<1$ in the proof of Theorem~\ref{thm:order_generic_completion} gives
\begin{equation}
 \tau_N(1-\eta)I_{M_N}\prec W_N\prec\tau_N(1+\eta)I_{M_N},
 \qquad
 \kappa_2(W_N)<\frac{1+\eta}{1-\eta},
\end{equation}
so the conditioning of the inverse symmetrizer is bounded uniformly in the state. The Frobenius normalization guarantees positive definiteness without an eigendecomposition. In implementation, $A_{i,N}W_N=S_{i,N}$ is solved as a linear system rather than forming $W_N^{-1}$ explicitly. For $N=3$, $M_N=10$, $p_N=6$, and $q_N=4$: six transport rows remain exact and each completion block is $4\times4$.
Figure~\ref{fig:hn_architecture} summarizes the flow from the resolved moments to the closed transport matrices. The networks act only on the resolved state, and every subsequent operation is a fixed algebraic map, so the structural properties hold for arbitrary network weights.

The directional outputs $\widehat K_x$ and $\widehat K_y$ determine the symmetric completion blocks $R_{i,N}$ in Eq.~\eqref{eq:general_SA_theorem}. This map also restricts the characteristic speeds by Eq.~\eqref{eq:speed_bounded_completion}; the hyperbolicity in Theorem~\ref{thm:order_generic_completion} does not depend on this restriction.

To interpret the learned rows, define, for $a+b=N$, the row vectors
\begin{align}
 h_{a+1,b}^{(x)}(\uu_N)
 &=\sqrt{\frac{2}{a+1}}\left[(A_{x,N})_{r_N(a,b),:}
       -\sqrt{\frac a2}\mathbf e_{r_N(a-1,b)}^{\mathsf T}\right],\\
 h_{a,b+1}^{(y)}(\uu_N)
 &=\sqrt{\frac{2}{b+1}}\left[(A_{y,N})_{r_N(a,b),:}
       -\sqrt{\frac b2}\mathbf e_{r_N(a,b-1)}^{\mathsf T}\right].
 \label{eq:2d_recover_rows}
\end{align}
The recurrence terms are omitted when $a=0$ or $b=0$, respectively. These vectors define predictions of the spatial derivatives $\partial_xm_{a+1,b}$ and $\partial_ym_{a,b+1}$ through $h_{a+1,b}^{(x)}\partial_x\uu_N$ and $h_{a,b+1}^{(y)}\partial_y\uu_N$. The row vectors themselves are not automatically gradients with respect to the resolved state. A shared scalar closure $g_{cd}$ for an interior unresolved moment, $c,d\ge1$ and $c+d=N+1$, would require
\begin{equation}
 h_{cd}^{(x)}=h_{cd}^{(y)}=(\nabla_{\uu_N}g_{cd})^{\mathsf T}.
 \label{eq:shared_consistency}
\end{equation}
In addition to agreement between directions, a continuously differentiable common row field $h_{cd}$ must satisfy the integrability conditions
\begin{equation}
 \frac{\partial(h_{cd})_j}{\partial u_k}
 =\frac{\partial(h_{cd})_k}{\partial u_j},\qquad 1\le j,k\le M_N,
\end{equation}
which are sufficient for a scalar potential on an open simply connected state domain. Symmetric completion alone does not enforce these conditions. The present method therefore evolves the matrices in Eq.~\eqref{eq:2d_closed_system} directly. The lower-order equations retain their known conservative fluxes, while the learned highest-order equations are interpreted in quasilinear form.

Regarding the training data, a DVM solution of Eq.~\eqref{eq:2d_bte} supplies the reference moments and labels. Let $\{\mathbf v_j,\omega_j\}_{j=1}^{N_v}$ denote velocity nodes and weights approximating integration over $\R^2$, and write $f_j(t,\mathbf x)\approx f(t,\mathbf x,\mathbf v_j)$. The velocity-discrete equations are
\begin{align}
 \partial_tf_j+v_{x,j}\partial_xf_j+v_{y,j}\partial_yf_j
 &=\sigma(M_j\rho_h-f_j),\\
 \rho_h:=\sum_{j=1}^{N_v}\omega_jf_j,\qquad
 M_j&:=\frac{M_2(\mathbf v_j)}{\sum_{\ell=1}^{N_v}\omega_\ell M_2(\mathbf v_\ell)}.
\end{align}
This normalization gives $\sum_j\omega_jM_j=1$. The collision update is
\begin{equation}
 f_j(t+\Delta t)=M_j\rho_h+e^{-\sigma\Delta t}(f_j(t)-M_j\rho_h).
 \label{eq:dvm_exact_collision}
\end{equation}
Reference moments through degree $N+1$ are computed using the same quadrature,
\begin{equation}
 m_{ab}^{\mathrm{DVM}}(t,\mathbf x)
 =\sum_{j=1}^{N_v}\omega_j f_j(t,\mathbf x)H_a(v_{x,j})H_b(v_{y,j}),
 \qquad a+b\le N+1.
 \label{eq:dvm_moment_postprocess}
\end{equation}
Spatial differentiation of the saved reference fields provides $\partial_x\uu_N$, $\partial_y\uu_N$, and the unresolved derivatives. For the smooth periodic tests, these labels are computed by Fourier differentiation.

To define the a priori objective, for each sample $s$, let $\mathbf d_i^{(s)}=(\partial_i\uu_N)^{\mathrm{DVM},(s)}$. For $a+b=N$, the supervised labels are the spatial derivatives of the unresolved moments themselves,
\begin{align}
 (\mathbf g_x^{(s)})_{a,b}&:=(\partial_xm_{a+1,b})^{\mathrm{DVM},(s)},\\
 (\mathbf g_y^{(s)})_{a,b}&:=(\partial_ym_{a,b+1})^{\mathrm{DVM},(s)}.
 \label{eq:unresolved_gradient_labels}
\end{align}
Both vectors lie in $\R^{q_N}$ and follow the ordering of the highest-degree layer. Using Eq.~\eqref{eq:2d_recover_rows}, their predictions are
\begin{align}
 (\widehat{\mathbf g}_x^{(s)})_{a,b}
 &:=h_{a+1,b}^{(x)}(\uu_N^{(s)})\mathbf d_x^{(s)},\\
 (\widehat{\mathbf g}_y^{(s)})_{a,b}
 &:=h_{a,b+1}^{(y)}(\uu_N^{(s)})\mathbf d_y^{(s)}.
 \label{eq:unresolved_gradient_predictions}
\end{align}
Thus the known derivative of the lower moment and the Hermite recurrence coefficient are removed from each learned row before evaluating the loss. For $N=3$, the eight supervised quantities are
\begin{align}
 \mathbf g_x&=(\partial_xm_{40},\partial_xm_{31},\partial_xm_{22},\partial_xm_{13})^{\mathsf T},\\
 \mathbf g_y&=(\partial_ym_{31},\partial_ym_{22},\partial_ym_{13},\partial_ym_{04})^{\mathsf T}.
 \label{eq:n3_gradient_targets}
\end{align}
The networks depend on $\uu_N$ only. The reference derivatives $\mathbf d_x,\mathbf d_y$ contract the learned rows in Eq.~\eqref{eq:unresolved_gradient_predictions}; they are not additional network inputs.

To balance components, trajectories, and the decay of their amplitudes, we normalize each derivative by its reference RMS within the corresponding trajectory and time snapshot. Let $\kappa_s$ and $t_s$ be the trajectory and the snapshot time of sample $s$, let $\mathcal S_\kappa$ collect the samples of trajectory $\kappa$, and let $\mathcal S_{\kappa,t}\subset\mathcal S_\kappa$ collect those with $t_s=t$. Define
\begin{align}
 r_{i,ab}(\kappa,t)&:=\left(\frac1{|\mathcal S_{\kappa,t}|}\sum_{s\in\mathcal S_{\kappa,t}}
          |(\mathbf g_i^{(s)})_{a,b}|^2\right)^{1/2},\\
 s_{i,ab}(\kappa,t)&:=\max\{r_{i,ab}(\kappa,t),\epsilon_{i,ab}(\kappa),\epsilon_0\},\qquad
 \epsilon_{i,ab}(\kappa):=\gamma\left(\frac1{|\mathcal S_\kappa|}
          \sum_{s\in\mathcal S_\kappa}|(\mathbf g_i^{(s)})_{a,b}|^2\right)^{1/2},
 \label{eq:gradient_snapshot_scales}
\end{align}
where $\gamma>0$ sets a relative floor and $\epsilon_0>0$ prevents a zero denominator. For a training set $\mathcal S_{\mathrm{tr}}$, the objective is
\begin{equation}
 \mathcal L_{\mathrm{cl}}(\phi)
 =\frac1{|\mathcal S_{\mathrm{tr}}|}
 \sum_{s\in\mathcal S_{\mathrm{tr}}}\sum_{i\in\{x,y\}}\sum_{a+b=N}
 \left[\frac{(\widehat{\mathbf g}_i^{(s)})_{a,b}-(\mathbf g_i^{(s)})_{a,b}}
 {s_{i,ab}(\kappa_s,t_s)}\right]^2.
 \label{eq:closure_training_loss}
\end{equation}
The trajectory $\kappa_s$ and time $t_s$ are used only to select the loss scale, so the learned constitutive coefficients remain functions of the resolved state. Since the trajectories differ in collision frequency and amplitude, a scale per trajectory prevents the trajectories with the largest gradients from dominating Eq.~\eqref{eq:closure_training_loss}. The RMS statistics and floors are computed over the full reference sample population of each trajectory before subsampling and before the split, and are reused for both training and validation.

For the two-dimensional DG experiments, the reference samples include the DVM grid points and Fourier-interpolated values and derivatives at element volume and face quadrature points. Let $\mathsf P_h^1$ denote elementwise projection onto the tensor Legendre space of degree one in each coordinate. Additional samples at the $3\times3$ volume quadrature points are the tuples
\begin{equation}
 \left(\mathsf P_h^1\uu_N,\ \partial_x\mathsf P_h^1\uu_N,\
 \partial_y\mathsf P_h^1\uu_N,\ \mathsf P_h^1\mathbf g_x,\ \mathsf P_h^1\mathbf g_y\right).
 \label{eq:projected_training_samples}
\end{equation}
In particular, the derivative labels are projected after differentiation, so that the samples cover the states at which the DG1 solver evaluates the closure. These samples are constructed from reference snapshots before training; no DG rollout enters Eq.~\eqref{eq:closure_training_loss}. The scalar values of the unresolved moments are needed for DVM postprocessing, but the moment solver uses the learned matrices directly.

Algorithm~\ref{alg:hn_training} summarizes the training procedure. The basis of $\mathcal L_N$, the reference samples, and the loss scales are computed once before the optimization loop, and each iteration evaluates only the networks and fixed algebraic maps.

\begin{algorithm}[htp!]
\caption{Training procedure for the HN closure}
\label{alg:hn_training}

\KwRequire{DVM reference snapshots $\{f_j(t_n,\mathbf x)\}$, truncation order $N$, parameters $\eta$, $\varepsilon$, $c_0$, $\delta_c$, $\gamma$, $\epsilon_0$.}
\KwEnsure{HN closure $\uu_N\mapsto\bigl(A_{x,N}(\uu_N),A_{y,N}(\uu_N)\bigr)$.}

\textcolor{gray}{$\triangleright$ Data preparation and initialization}\;
Construct $F_x^{(N)}$, $F_y^{(N)}$ and a basis $\{B_k^{(N)}\}_{k=1}^{d_N}$ of $\mathcal L_N=\ker\mathcal C_N$
\graycomment{By Theorem~\ref{thm:order_generic_completion}}\;
Compute $\uu_N$, $\partial_x\uu_N$, $\partial_y\uu_N$, $\mathbf g_x$, and $\mathbf g_y$ at grid, quadrature, and projected sample points
\graycomment{By Eqs.~\eqref{eq:dvm_moment_postprocess}, \eqref{eq:unresolved_gradient_labels}, \eqref{eq:projected_training_samples}}\;
Compute the loss scales $s_{i,ab}(\kappa,t)$ and the input statistics $\mu_j$, $s_j^{\mathrm{in}}$
\graycomment{By Eq.~\eqref{eq:gradient_snapshot_scales}}\;
Split the samples into training and validation sets, and initialize the network parameters $\phi$\;

\While{stopping criterion is not satisfied}{
    Sample a mini batch $\mathcal B\subset\mathcal S_{\mathrm{tr}}$\;

    Evaluate $\theta_k^{(N)}$, $\widehat\tau_N$, $\widehat K_x$, and $\widehat K_y$ at the standardized inputs for all samples in $\mathcal B$\;

    Form $E_N=\widetilde W_N/(1+\|\widetilde W_N\|_F)$ and
    $W_N=\tau_N(I_{M_N}+\eta E_N)$
    \graycomment{By Eq.~\eqref{eq:2d_Wspd}; $W_N\succ0$ by construction}\;

    Compute the Cholesky factor $W_N=LL^{\mathsf T}$ and the known blocks $[E_i\ G_i]$ for $i\in\{x,y\}$
    \graycomment{By Eq.~\eqref{eq:bounded_similarity}}\;

    Set $c_i=\max\{c_0,(1+\delta_c)\|[E_i\ G_i]\|_2\}$ and form $J_i$ and $B_i$
    \graycomment{By Eq.~\eqref{eq:speed_bounded_completion}}\;

    Recover $S_{i,N}=LB_iL^{\mathsf T}$ and solve $A_{i,N}W_N=S_{i,N}$ for $A_{i,N}$\;

    Extract the learned rows $h_{a+1,b}^{(x)}$ and $h_{a,b+1}^{(y)}$ for $a+b=N$
    \graycomment{By Eq.~\eqref{eq:2d_recover_rows}}\;

    Predict $\widehat{\mathbf g}_x=h^{(x)}\mathbf d_x$ and $\widehat{\mathbf g}_y=h^{(y)}\mathbf d_y$
    \graycomment{By Eq.~\eqref{eq:unresolved_gradient_predictions}}\;

    Compute $\mathcal L_{\mathrm{cl}}$ over $\mathcal B$ and update $\phi$ by a gradient step
    \graycomment{By Eq.~\eqref{eq:closure_training_loss}}\;
}

\Return $A_{x,N}(\cdot)$ and $A_{y,N}(\cdot)$\;
\end{algorithm}

\subsection{Discontinuous Galerkin discretization}
\label{sec:jacobian_first_numerics}

The fixed lower-order rows in Eq.~\eqref{eq:general_fixed_rows} permit conservative fluxes, while the learned highest-order rows require a nonconservative treatment. We combine these two forms in a modal DG discretization. Let $\Pi_L$ and $\Pi_H$ extract the first $p_N$ and last $q_N$ components of a vector in $\R^{M_N}$, respectively. Define
\begin{equation}
 \mathbf f_i^L(\uu):=F_i^{(N)}\uu,\qquad
 \mathbf Q(\uu):=S_\sigma\uu,\qquad i\in\{x,y\}.
 \label{eq:dg_lower_flux}
\end{equation}
With $\uu_L=\Pi_L\uu_N$ and $\uu_H=\Pi_H\uu_N$, the system becomes
\begin{align}
 \partial_t\uu_L+\partial_x\mathbf f_x^L(\uu_N)
                         +\partial_y\mathbf f_y^L(\uu_N)&=\Pi_L\mathbf Q(\uu_N),\\
 \partial_t\uu_H+\Pi_H\left[A_{x,N}(\uu_N)\partial_x\uu_N
                         +A_{y,N}(\uu_N)\partial_y\uu_N\right]&=\Pi_H\mathbf Q(\uu_N).
 \label{eq:dg_partitioned_system}
\end{align}
Thus the conservative block contains all moments of degree less than $N$, and the nonconservative block contains the degree-$N$ moment equations. For $N=3$, these blocks contain six and four equations, respectively. Both blocks evolve the same resolved state.

For the element approximation, let $\mathcal T_h$ be a rectangular mesh of the periodic domain $\Omega$. For $K\in\mathcal T_h$, use the affine map
\begin{equation}
 x=x_K+\frac{h_x^K}{2}\xi,\qquad y=y_K+\frac{h_y^K}{2}\eta,
 \qquad (\xi,\eta)\in[-1,1]^2.
\end{equation}
For spatial polynomial degree $p$, let $P_j$ denote the Legendre polynomial of degree $j$ and set
\begin{equation}
 V_h^K:=\operatorname{span}\{\psi_{ab}(\xi,\eta)=P_a(\xi)P_b(\eta):0\le a,b\le p\},
 \qquad
 \uu_h|_K=\sum_{a,b=0}^{p}\mathbf U_{ab}^K(t)\psi_{ab}(\xi,\eta).
 \label{eq:dg_modal_expansion}
\end{equation}
The spatial degree $p$ is independent of the Hermite truncation order $N$. In particular, DG1 uses the four tensor modes $1,\xi,\eta,\xi\eta$ on each element. The Legendre mass matrix is diagonal:
\begin{equation}
 (\mathbf M_K)_{(i,j),(a,b)}
 =\int_K\psi_{ij}\psi_{ab}\,d\mathbf x
 =\frac{h_x^Kh_y^K}{(2i+1)(2j+1)}\delta_{ia}\delta_{jb}.
 \label{eq:dg_mass_matrix}
\end{equation}
Each matrix entry acts separately on the $M_N$ moment components. Ordering modes with the second index varying fastest also gives
\begin{align}
 \mathbf M_K&=\frac{h_x^Kh_y^K}{4}(\mathbf M^x\otimes\mathbf M^y),\\
 \mathbf N_{x,K}&=\frac{h_y^K}{2}(\mathbf S^x\otimes\mathbf M^y),\qquad
 \mathbf N_{y,K}=\frac{h_x^K}{2}(\mathbf M^x\otimes\mathbf S^y).
 \label{eq:dg_element_matrices}
\end{align}
where $M^x_{ia}=\int_{-1}^1P_iP_a\,d\xi$, $S^x_{ia}=\int_{-1}^1P_i'P_a\,d\xi$, with analogous definitions in $y$. The matrices $\mathbf N_{i,K}$ represent testing the conservative flux against a derivative of the basis.

Regarding the interface treatment, consider a face of an element $K$, and let $\mathbf n=(n_x,n_y)$ be the outward unit normal, and let $\uu^-$ and $\uu^+$ denote the interior and exterior traces. Periodic faces use the trace of the neighboring periodic element. Define
\begin{align}
 A_{\mathbf n}(\uu)&:=n_xA_{x,N}(\uu)+n_yA_{y,N}(\uu),\\
 \Psi(s;\uu^-,\uu^+)&:=\uu^-+s\Delta\uu,\qquad
 \Delta\uu:=\uu^+-\uu^-,\quad 0\le s\le1,\\
 \mathcal P_{\mathbf n}(\uu^-,\uu^+)
 &:=\int_0^1A_{\mathbf n}(\Psi(s))\Delta\uu\,ds.
 \label{eq:dg_path_integral}
\end{align}
The path integral specifies the contribution of the nonconservative product across a jump, following the path treatment of learned moment systems in \cite{ChristliebEtAl2024}. It is an interface term and does not reconstruct scalar unresolved moments. With four Gauss--Legendre nodes $s_r$ and weights $\omega_r$ on $[0,1]$, use
\begin{align}
 \mathcal P_{\mathbf n}^{Q}&:=\sum_{r=1}^4\omega_r A_{\mathbf n}(\Psi(s_r))\Delta\uu,\\
 \mathcal D_{\mathbf n}^{\pm}(\uu^-,\uu^+)
 &:=\frac12\left[\mathcal P_{\mathbf n}^{Q}\pm\alpha_e\Delta\uu\right].
 \label{eq:dg_path_rusanov}
\end{align}
Consequently $\mathcal D_{\mathbf n}^-+\mathcal D_{\mathbf n}^+=\mathcal P_{\mathbf n}^{Q}$, and both fluctuations vanish for equal traces. The element whose outward normal is $\mathbf n$ receives $\mathcal D_{\mathbf n}^-$. The neighboring element uses reversed traces and normal, giving $\mathcal D_{-\mathbf n}^-(\uu^+,\uu^-)=\mathcal D_{\mathbf n}^+(\uu^-,\uu^+)$.

The dissipation coefficient is computed from the symmetric matrix in Eq.~\eqref{eq:bounded_similarity}. For a unit normal, set
\begin{equation}
 B_{\mathbf n}(\uu):=L^{-1}(n_xS_{x,N}+n_yS_{y,N})L^{-\mathsf T},\qquad
 b_{\mathbf n}(\uu):=\rho\bigl(B_{\mathbf n}(\uu)\bigr)=\rho\bigl(A_{\mathbf n}(\uu)\bigr),
 \label{eq:dg_speed_bound}
\end{equation}
where $\rho$ denotes the spectral radius. Since $B_{\mathbf n}$ is symmetric and similar to $A_{\mathbf n}$, $b_{\mathbf n}$ is computed by a symmetric eigenvalue solver and is the exact maximum characteristic speed; by the construction in Eq.~\eqref{eq:speed_bounded_completion}, $b_{\mathbf n}\le c$ in the coordinate directions for every network output. At each face quadrature point, choose
\begin{equation}
 \alpha_e:=\max_{s\in\{0,s_1,s_2,s_3,s_4,1\}}
                  b_{\mathbf n}(\Psi(s)).
 \label{eq:dg_face_speed}
\end{equation}
It is the largest characteristic speed at the evaluated path states.

For the element equations, take any scalar test function $v_h\in V_h^K$. Integration by parts in the lower block of Eq.~\eqref{eq:dg_partitioned_system} gives
\begin{align}
 \int_Kv_h\partial_t(\Pi_L\uu_h)\,d\mathbf x
 &-\sum_{i\in\{x,y\}}\int_K(\partial_i v_h)\mathbf f_i^L(\uu_h)\,d\mathbf x
 +\int_{\partial K}v_h^-\widehat{\mathbf f}_{\mathbf n}^L\,ds\nonumber\\
 &=\int_Kv_h\Pi_L\mathbf Q(\uu_h)\,d\mathbf x,
 \label{eq:dg_conservative_weak}
\end{align}
with the common Rusanov flux
\begin{equation}
 \widehat{\mathbf f}_{\mathbf n}^L(\uu^-,\uu^+)
 :=\frac12F_{\mathbf n}^{(N)}(\uu^-+\uu^+)
     -\frac{\alpha_e}{2}\Pi_L(\uu^+-\uu^-),\qquad
 F_{\mathbf n}^{(N)}:=n_xF_x^{(N)}+n_yF_y^{(N)}.
 \label{eq:dg_lower_rusanov}
\end{equation}
For the highest-order block, retain the strong volume product and add the incoming path fluctuation:
\begin{align}
 \int_Kv_h\partial_t(\Pi_H\uu_h)\,d\mathbf x
 &+\sum_{i\in\{x,y\}}\int_Kv_h\Pi_H\left[A_{i,N}(\uu_h)\partial_i\uu_h\right]d\mathbf x\nonumber\\
 &+\int_{\partial K}v_h^-\Pi_H\mathcal D_{\mathbf n}^-(\uu_h^-,\uu_h^+)\,ds
 =\int_Kv_h\Pi_H\mathbf Q(\uu_h)\,d\mathbf x.
 \label{eq:dg_nonconservative_weak}
\end{align}
The volume term accounts for variation within an element; the fluctuation accounts for its interface jump. Both evaluate the same learned matrices at their respective local states.

These two element forms agree on every conservative row. Indeed, the fixed rows give the exact identity
\begin{equation}
 \Pi_L\mathcal P_{\mathbf n}^{Q}
 =F_{\mathbf n}^{(N)}(\uu^+-\uu^-),\qquad
 \Pi_L\mathcal D_{\mathbf n}^-
 =\widehat{\mathbf f}_{\mathbf n}^L-\mathbf f_{\mathbf n}^L(\uu^-),
 \label{eq:dg_conservative_reduction}
\end{equation}
where $\mathbf f_{\mathbf n}^L=n_x\mathbf f_x^L+n_y\mathbf f_y^L$. Applying integration by parts to the corresponding strong volume term therefore recovers Eq.~\eqref{eq:dg_conservative_weak}. The classical $P_N$ closure sets every unresolved moment of total degree $N+1$ to zero. All its transport rows are constant, so the entire discretization reduces to conservative DG. Taking $v_h=1$ in the lower block makes the fluxes on interior faces cancel between neighboring elements. The periodic density integral is consequently conserved, since its collision source vanishes.

Taking $v_h=\psi_{ij}$ in Eqs.~\eqref{eq:dg_conservative_weak}--\eqref{eq:dg_nonconservative_weak} and collecting modal coefficients yields
\begin{equation}
 \mathbf M_K\frac{d\mathbf U_K}{dt}
 =\mathbf R_K^{\mathrm{vol}}(\mathbf U_K)
  +\mathbf R_K^{\mathrm{face}}(\mathbf U_K,\{\mathbf U_{K'}\}_{K'\sim K})
  +\mathbf R_K^{\mathrm{src}}(\mathbf U_K).
 \label{eq:dg_element_semidiscrete}
\end{equation}
Here $K'\sim K$ denotes a face neighbor, and the residual entries for a mode $\psi_{ij}$ are
\begin{align}
 (\mathbf R_K^{\mathrm{vol}})_{ij}
 &=\begin{bmatrix}
 \displaystyle\sum_{\ell\in\{x,y\}}\int_K(\partial_\ell\psi_{ij})\mathbf f_\ell^L(\uu_h)\,d\mathbf x\\[2pt]
 \displaystyle-\sum_{\ell\in\{x,y\}}\int_K\psi_{ij}\Pi_H[A_{\ell,N}(\uu_h)\partial_\ell\uu_h]\,d\mathbf x
 \end{bmatrix},\\
 (\mathbf R_K^{\mathrm{face}})_{ij}
 &=-\int_{\partial K}\psi_{ij}^-\begin{bmatrix}
 \widehat{\mathbf f}_{\mathbf n}^L\\ \Pi_H\mathcal D_{\mathbf n}^-
 \end{bmatrix}ds,\qquad
 (\mathbf R_K^{\mathrm{src}})_{ij}=\int_K\psi_{ij}S_\sigma\uu_h\,d\mathbf x.
 \label{eq:dg_modal_residuals}
\end{align}
Volume integrals use tensor Gauss--Legendre quadrature with $\max\{p+2,3\}$ nodes per coordinate, and face integrals use the same number of nodes tangentially. The linear conservative volume terms and the mass matrix are integrated exactly; nonlinear products and path integrals are evaluated by the stated quadrature rules. Initial modal coefficients are obtained by elementwise $L^2$ projection of the Fourier-interpolated DVM initial field. DVM comparisons of cell averages use the same element averaging, rather than comparing a DG average with a kinetic point value.

\subsection{Time integration}
\label{sec:time_integration}

After assembling Eq.~\eqref{eq:dg_element_semidiscrete}, let $\mathbf U$ collect all element coefficients and define the transport operator $\mathcal L_h(\mathbf U)$ by applying each $\mathbf M_K^{-1}$ to its volume and face residuals. A transport step starting from $\mathbf V^{(0)}$ uses the third-order strong-stability-preserving Runge--Kutta method \cite{ShuOsher1988},
\begin{align}
    \mathbf V^{(1)} &= \mathbf V^{(0)}+\Delta t\,\mathcal L_h(\mathbf V^{(0)}),\\
    \mathbf V^{(2)} &= \frac34\mathbf V^{(0)}+\frac14\left[\mathbf V^{(1)}+\Delta t\,\mathcal L_h(\mathbf V^{(1)})\right],\\
    \mathbf V^{(3)}&=\frac13\mathbf V^{(0)}+\frac23\left[\mathbf V^{(2)}+\Delta t\,\mathcal L_h(\mathbf V^{(2)})\right].
    \label{eq:ssprk3}
\end{align}
Set $\mathcal T_{\Delta t}(\mathbf V^{(0)}):=\mathbf V^{(3)}$. The learned matrices, path integrals, and dissipation speeds are evaluated at the current stage state. Both moment blocks participate in every stage.

For the constant collision frequency considered here, the source update is exact. On every element and for every spatial mode, define
\begin{equation}
 \mathcal C_{\Delta t}:\quad
 U_{ij,00}^K\mapsto U_{ij,00}^K,\qquad
 U_{ij,ab}^K\mapsto e^{-\sigma\Delta t}U_{ij,ab}^K,
 \quad 1\le a+b\le N.
    \label{eq:moment_exact_collision}
\end{equation}
Here $i,j$ index spatial modes and $a,b$ index Hermite moments. Combining this update with Eq.~\eqref{eq:ssprk3} gives the Strang composition
\begin{equation}
    \mathbf U^{n+1}=
    \mathcal C_{\Delta t/2}
    \mathcal T_{\Delta t}
    \mathcal C_{\Delta t/2}\mathbf U^n.
    \label{eq:strang_moment_update}
\end{equation}
Thus the transport integrator is third order, while the full transport--collision splitting is second order in time.

On a uniform rectangular mesh, let $\beta_x,\beta_y$ be the maxima of the bounds in Eq.~\eqref{eq:dg_speed_bound} over the volume, face, and path states evaluated in the coordinate directions. The time step satisfies
\begin{equation}
 \Delta t\le\frac{C_{\mathrm{CFL}}}
 {(2p+1)(\beta_x/h_x+\beta_y/h_y)}.
 \label{eq:dg_cfl}
\end{equation}
The factor $2p+1$ accounts for the spatial polynomial degree. The step size is chosen from the speeds of the previous step with a $2\%$ margin. If a Runge--Kutta stage violates Eq.~\eqref{eq:dg_cfl}, the step is repeated with $\Delta t$ equal to $95\%$ of the bound given by that stage. Steps are also shortened to reach the prescribed output times. All reported DG1 calculations use $C_{\mathrm{CFL}}=0.15$.

Algorithm~\ref{alg:hn_dg} summarizes the resulting moment solver. The closure is evaluated at every Runge--Kutta stage at the volume quadrature points, the face traces, and the path nodes. No reference data or retraining enter this procedure.

\begin{algorithm}[htp!]
\caption{DG computational procedure with the HN closure}
\label{alg:hn_dg}

\KwRequire{Mesh $\mathcal T_h$, polynomial degree $p$, collision frequency $\sigma$, initial moments $\uu_N(\mathbf x,0)$, final time $T$, $C_{\mathrm{CFL}}$, and trained HN closure.}
\KwEnsure{DG moment solution $\uu_h(\cdot,T)$.}

Initialize $\mathbf U_K$ by elementwise $L^2$ projection of $\uu_N(\cdot,0)$ for all $K\in\mathcal T_h$; set $t=0$\;

\While{$t<T$}{
    Choose $\Delta t$ from the characteristic speeds of the previous step and shorten it to reach the next output time
    \graycomment{By Eq.~\eqref{eq:dg_cfl}}\;

    Apply the half collision step $\mathbf U\leftarrow\mathcal C_{\Delta t/2}\mathbf U$
    \graycomment{By Eq.~\eqref{eq:moment_exact_collision}}\;

    \For{each SSPRK3 stage}{
        Evaluate $\uu_h$ and $\partial_x\uu_h$, $\partial_y\uu_h$ at the volume quadrature points of every $K$\;

        Evaluate $A_{x,N}(\uu_h)$ and $A_{y,N}(\uu_h)$ with the HN closure, and assemble $\mathbf R_K^{\mathrm{vol}}$
        \graycomment{By Eq.~\eqref{eq:dg_modal_residuals}}\;

        Initialize $\mathbf R_K^{\mathrm{face}}=\mathbf 0$\;

        \ForEach{face $e=K^-\cap K^+$ with unit normal $\mathbf n$ outward from $K^-$}{
            Evaluate the traces $\uu^-$ and $\uu^+$ at the face quadrature points\;

            Evaluate $A_{\mathbf n}$ and $b_{\mathbf n}$ at the path states $\Psi(s_r;\uu^-,\uu^+)$, $r=1,\dots,4$, and at $\uu^\pm$
            \graycomment{Real spectrum by construction}\;

            Set $\alpha_e=\max_s b_{\mathbf n}(\Psi(s))$ and form $\mathcal P_{\mathbf n}^Q$
            \graycomment{By Eqs.~\eqref{eq:dg_path_rusanov}--\eqref{eq:dg_face_speed}}\;

            Form the conservative flux $\widehat{\mathbf f}_{\mathbf n}^L$ and the fluctuations $\mathcal D_{\mathbf n}^{\pm}$
            \graycomment{By Eqs.~\eqref{eq:dg_path_rusanov}, \eqref{eq:dg_lower_rusanov}}\;

            Add $-[\widehat{\mathbf f}_{\mathbf n}^L;\Pi_H\mathcal D_{\mathbf n}^-]$ to $\mathbf R_{K^-}^{\mathrm{face}}$ and
            $-[-\widehat{\mathbf f}_{\mathbf n}^L;\Pi_H\mathcal D_{\mathbf n}^+]$ to $\mathbf R_{K^+}^{\mathrm{face}}$, tested against the face basis\;
        }

        Update the stage by $\mathcal L_h(\mathbf U)=\mathbf M_K^{-1}(\mathbf R_K^{\mathrm{vol}}+\mathbf R_K^{\mathrm{face}})$
        \graycomment{By Eq.~\eqref{eq:ssprk3}}\;

        \If{the stage state violates Eq.~\eqref{eq:dg_cfl}}{
            Reduce $\Delta t$ and restart the transport step\;
        }
    }

    Apply the half collision step $\mathbf U\leftarrow\mathcal C_{\Delta t/2}\mathbf U$ and set $t\leftarrow t+\Delta t$\;
}

\Return $\uu_h(\cdot,T)$\;
\end{algorithm}

\section{Numerical experiments}
\label{Sec:numerical_experiments}

In this section, we validate the proposed hyperbolic neural (HN) closure for the two-dimensional linear Boltzmann equation by DG simulations of the closed moment system.
The three experiments share the kinetic equation~\eqref{eq:2d_bte}, the collision operator~\eqref{eq:2d_linear_collision}, and the periodic domain $\Omega=[0,1]^2$. They differ in the initial distribution and in the collision frequency $\sigma$, so that each experiment probes a different collisional regime:
\begin{itemize}
    \item Experiment~\ref{ex1_2d}: a non-equilibrium Maxwellian mixture in the transition regime, $\sigma=2$;
    \item Experiment~\ref{ex2_2d}: a localized Gaussian density pulse in the scattering-dominated regime, $\sigma=10$;
    \item Experiment~\ref{ex3_2d}: a converging ring-shaped density pulse in the scattering-dominated regime, $\sigma=8$.
\end{itemize}
Since $\Omega$ has unit length and the equilibrium velocity scale is of order one, $\mathrm{Kn}:=\sigma^{-1}$ serves as a Knudsen number. The learned moment system is evolved by the partially conservative DG method in Section~\ref{sec:jacobian_first_numerics}, with SSPRK3 transport updates and exact collision steps combined by Eq.~\eqref{eq:strang_moment_update}. Kinetic DVM solutions supply the reference moments and spatial derivative labels. In every experiment, the HN closure is compared with the classical $P_3$, $P_7$, and $P_{15}$ truncations, which evolve $10$, $36$, and $136$ moments, respectively. These truncations have constant symmetric transport matrices and are solved by the same DG method with the same mesh, polynomial degree, and CFL number, so the comparison isolates the effect of the closure.

For all three experiments, we use $N=3$, so that the resolved state is
\begin{equation}
\uu_3=\begin{bmatrix}m_{00}&m_{10}&m_{01}&m_{20}&m_{11}&m_{02}&m_{30}&m_{21}&m_{12}&m_{03}\end{bmatrix}^{\mathsf T}\in\R^{10},
\label{eq:exp_state_n3}
\end{equation}
and the unresolved moments are $m_{40},m_{31},m_{22},m_{13},m_{04}$. From Eq.~\eqref{eq:2d_moment_eq}, the six equations of degree less than three are
\[
\frac{\partial m_{00}}{\partial t}
+\frac{1}{\sqrt2}\frac{\partial m_{10}}{\partial x}
+\frac{1}{\sqrt2}\frac{\partial m_{01}}{\partial y}=0,
\]
\[
\frac{\partial m_{10}}{\partial t}
+\frac{\partial}{\partial x}\left(\frac{1}{\sqrt2}m_{00}+m_{20}\right)
+\frac{1}{\sqrt2}\frac{\partial m_{11}}{\partial y}=-\sigma m_{10},
\]
\[
\frac{\partial m_{01}}{\partial t}
+\frac{1}{\sqrt2}\frac{\partial m_{11}}{\partial x}
+\frac{\partial}{\partial y}\left(\frac{1}{\sqrt2}m_{00}+m_{02}\right)=-\sigma m_{01},
\]
\[
\frac{\partial m_{20}}{\partial t}
+\frac{\partial}{\partial x}\left(m_{10}+\sqrt{\tfrac32}\,m_{30}\right)
+\frac{1}{\sqrt2}\frac{\partial m_{21}}{\partial y}=-\sigma m_{20},
\]
\[
\frac{\partial m_{11}}{\partial t}
+\frac{\partial}{\partial x}\left(\frac{1}{\sqrt2}m_{01}+m_{21}\right)
+\frac{\partial}{\partial y}\left(\frac{1}{\sqrt2}m_{10}+m_{12}\right)=-\sigma m_{11},
\]
\[
\frac{\partial m_{02}}{\partial t}
+\frac{1}{\sqrt2}\frac{\partial m_{12}}{\partial x}
+\frac{\partial}{\partial y}\left(m_{01}+\sqrt{\tfrac32}\,m_{03}\right)=-\sigma m_{02}.
\]
These equations involve only resolved moments and are discretized in conservative form. The four highest-degree equations are
\[
\frac{\partial m_{30}}{\partial t}
+\sqrt{\tfrac32}\frac{\partial m_{20}}{\partial x}
+\sqrt2\,\frac{\partial m_{40}}{\partial x}
+\frac{1}{\sqrt2}\frac{\partial m_{31}}{\partial y}
=-\sigma m_{30},
\]
\[
\frac{\partial m_{21}}{\partial t}
+\frac{\partial m_{11}}{\partial x}
+\sqrt{\tfrac32}\,\frac{\partial m_{31}}{\partial x}
+\frac{1}{\sqrt2}\frac{\partial m_{20}}{\partial y}
+\frac{\partial m_{22}}{\partial y}
=-\sigma m_{21},
\]
\[
\frac{\partial m_{12}}{\partial t}
+\frac{1}{\sqrt2}\frac{\partial m_{02}}{\partial x}
+\frac{\partial m_{22}}{\partial x}
+\frac{\partial m_{11}}{\partial y}
+\sqrt{\tfrac32}\,\frac{\partial m_{13}}{\partial y}
=-\sigma m_{12},
\]
\[
\frac{\partial m_{03}}{\partial t}
+\frac{1}{\sqrt2}\frac{\partial m_{13}}{\partial x}
+\sqrt{\tfrac32}\frac{\partial m_{02}}{\partial y}
+\sqrt2\,\frac{\partial m_{04}}{\partial y}
=-\sigma m_{03},
\]
where $\partial_xm_{40}$, $\partial_xm_{31}$, $\partial_xm_{22}$, $\partial_xm_{13}$, $\partial_ym_{31}$, $\partial_ym_{22}$, $\partial_ym_{13}$, and $\partial_ym_{04}$ are the unresolved derivatives in Eq.~\eqref{eq:n3_gradient_targets}. The HN closure replaces them by the learned rows in Eq.~\eqref{eq:2d_recover_rows},
\begin{equation}
\frac{\partial m_{a+1,b}}{\partial x}\approx h_{a+1,b}^{(x)}(\uu_3)\frac{\partial\uu_3}{\partial x},
\qquad
\frac{\partial m_{a,b+1}}{\partial y}\approx h_{a,b+1}^{(y)}(\uu_3)\frac{\partial\uu_3}{\partial y},
\qquad a+b=3.
\label{eq:exp_closure_substitution}
\end{equation}
Equivalently, the closed system is
\begin{equation}
\frac{\partial\uu_3}{\partial t}
+A_{x,3}(\uu_3)\frac{\partial\uu_3}{\partial x}
+A_{y,3}(\uu_3)\frac{\partial\uu_3}{\partial y}
=S_\sigma\uu_3,
\qquad
S_\sigma=\operatorname{diag}(0,-\sigma,\ldots,-\sigma),
\label{eq:exp_closed_system}
\end{equation}
where the first six rows of $A_{x,3}$ and $A_{y,3}$ are the fixed Hermite rows in Eq.~\eqref{eq:general_fixed_rows}, and the last four rows are given by the symmetric completion in Theorem~\ref{thm:order_generic_completion}, with the bounded completion in Eq.~\eqref{eq:speed_bounded_completion}. The classical $P_3$ truncation sets all of these unresolved terms to zero.

For the reference solutions, the DVM in every experiment discretizes the transport of each velocity node by the DG method of Section~\ref{sec:jacobian_first_numerics} with $p=1$ and the upwind flux on the $50\times50$ mesh, and combines SSPRK3 transport steps with Eq.~\eqref{eq:dvm_exact_collision} in a Strang composition, with $C_{\mathrm{CFL}}=0.15$ in Eq.~\eqref{eq:dg_cfl}. The velocity nodes form a polar product grid. Writing $\mathbf v=r(\cos\theta,\sin\theta)$, any integrable $\varphi$ satisfies
\[
\int_{\R^2}\varphi(\mathbf v)M_2(\mathbf v)\,d\mathbf v
=\frac1{2\pi}\int_0^{2\pi}\int_0^\infty\varphi\bigl(r\cos\theta,r\sin\theta\bigr)\,2re^{-r^2}\,dr\,d\theta.
\]
Let $r_\ell,w_\ell^{\mathrm{GL}}$, $\ell=1,\ldots,N_r$, be the Gauss--Legendre nodes and weights on $[0,r_{\max}]$, and let $\theta_k=2\pi(k-\tfrac12)/N_\theta$, $k=1,\ldots,N_\theta$. The unweighted integration convention in Eq.~\eqref{eq:dvm_moment_postprocess} then uses
\begin{equation}
 \mathbf v_{\ell k}=r_\ell(\cos\theta_k,\sin\theta_k),\qquad
 \omega_{\ell k}M_2(\mathbf v_{\ell k})=\frac{\widehat w_\ell}{N_\theta},\qquad
 \widehat w_\ell=\frac{w_\ell^{\mathrm{GL}}\,r_\ell e^{-r_\ell^2}}{\sum_{m=1}^{N_r}w_m^{\mathrm{GL}}\,r_m e^{-r_m^2}},
 \label{eq:polar_velocity_quadrature}
\end{equation}
so that $\sum_{\ell,k}\omega_{\ell k}M_2(\mathbf v_{\ell k})=1$, consistent with the normalization of $M_j$. We use $r_{\max}=8$, $N_\theta=128$, and $N_r=96$, that is, $N_v=12{,}288$ velocity nodes. With these values, the normalized Hermite polynomials of degree at most $16$, which suffice for the $P_{15}$ comparison, are orthonormal under Eq.~\eqref{eq:polar_velocity_quadrature} to within $10^{-9}$.

The polar grid avoids the ray effects of tensor velocity grids. For density, we distinguish the relative error of the full field and the error relative to the reference density variation,
\begin{equation}
 E_\rho:=\frac{\|\rho_h-\rho_{\mathrm{DVM}}\|_{2,h}}{\|\rho_{\mathrm{DVM}}\|_{2,h}},\qquad
 E_{\delta\rho}:=\frac{\|\rho_h-\rho_{\mathrm{DVM}}\|_{2,h}}
 {\|\rho_{\mathrm{DVM}}-\langle\rho_{\mathrm{DVM}}\rangle\|_{2,h}}.
 \label{eq:density_error_metrics}
\end{equation}
Here both fields are cell averages, $\langle\cdot\rangle$ is the spatial mean, and $\|z\|_{2,h}^2=\sum_K|K|\,|z_K|^2$. Since all three initial densities are small perturbations of a uniform state that decay in time, $E_{\delta\rho}$ is the more stringent measure.

Regarding the computational settings, Table~\ref{tab:exp_settings} lists the DVM and DG parameters of the three experiments. In each experiment, the solutions are compared at $t=0.1$, $0.2$, $0.3$, $0.4$, and $0.5$, and the errors in Eq.~\eqref{eq:density_error_metrics} are reported for the HN closure and for the $P_3$, $P_7$, and $P_{15}$ truncations. The DG solutions, the DVM references, and the network evaluations run on a single NVIDIA GeForce RTX 4090 GPU; the wall-clock times in the tables are measured on this hardware. The DVM reference requires $881$\,s in each experiment.

\begin{table}[htbp!]
\footnotesize
\renewcommand{\arraystretch}{1.12}
\setlength{\tabcolsep}{5pt}
\centering
\caption{Kinetic reference and DG settings of the three experiments.}
\label{tab:exp_settings}
\begin{tabular}{llll}
\hline
 & Experiment~\ref{ex1_2d} & Experiment~\ref{ex2_2d} & Experiment~\ref{ex3_2d} \\
\hline
Collision frequency $\sigma$ & $2$ & $10$ & $8$ \\
Knudsen number $\mathrm{Kn}=\sigma^{-1}$ & $0.5$ & $0.1$ & $0.125$ \\
Initial distribution & Eq.~\eqref{eq:2d_mixture_initial} & Eq.~\eqref{eq:2d_blob_test} & Eq.~\eqref{eq:2d_ring_test} \\
\hline
DVM spatial grid & $50\times50$ & $50\times50$ & $50\times50$ \\
Velocity nodes $N_\theta\times N_r$ & $128\times96$ & $128\times96$ & $128\times96$ \\
DVM time step $\Delta t_{\mathrm{DVM}}$ & $6.1\times10^{-5}$ & $6.1\times10^{-5}$ & $6.1\times10^{-5}$ \\
\hline
DG mesh & $50\times50$ & $50\times50$ & $50\times50$ \\
Polynomial degree $p$ & $1$ & $1$ & $1$ \\
Hermite order of HN closure & $N=3$ & $N=3$ & $N=3$ \\
Classical truncations & $P_3,P_7,P_{15}$ & $P_3,P_7,P_{15}$ & $P_3,P_7,P_{15}$ \\
CFL number $C_{\mathrm{CFL}}$ & $0.15$ & $0.15$ & $0.15$ \\
Final time $T$ & $0.5$ & $0.5$ & $0.5$ \\
\hline
\end{tabular}
\end{table}

\subsection{Training of the HN closure}
\label{sec:exp_training}

A single HN closure is trained for all three experiments, following~\cite{ChristliebEtAl2024}. The training data consist of the DVM trajectories of the three experiments and of trajectories from three randomized families of initial distributions, one for each experiment (Table~\ref{tab:exp_ic_families}), with collision frequencies sampled as
\begin{equation}
\log_{10}\sigma\sim\mathcal U\bigl(\log_{10}0.5,\log_{10}20\bigr).
\label{eq:exp_sigma_sampling}
\end{equation}
Family~I contains $32$ trajectories and Families~II and~III contain $16$ each, so that, together with the three experiment trajectories, the data consist of $67$ trajectories. Snapshots at $t=0,0.02,\ldots,0.5$ are used, and the DG simulations are run to $T=0.5$. The experiments therefore do not test extrapolation to unseen data; they test whether a single closure trained a priori, with guaranteed hyperbolicity, represents the three cases and remains accurate and stable inside the DG solver. The samples of Section~\ref{sec:neural_parameterization} are generated on the grid of each DVM trajectory. After the loss scales in Eq.~\eqref{eq:gradient_snapshot_scales} are computed, $3\times10^5$ samples are drawn from each trajectory, and the pooled samples are split at random into $80\%$ for training and $20\%$ for testing.

\begin{table}[htbp!]
\footnotesize
\renewcommand{\arraystretch}{1.15}
\setlength{\tabcolsep}{5pt}
\centering
\caption{Randomized families of initial distributions used to train the single HN closure. Each parameter is sampled uniformly from the listed range, and $\sigma$ is sampled from Eq.~\eqref{eq:exp_sigma_sampling}. The last column lists the parameters of the corresponding experiment, whose trajectory is also included in the training data. The ranges of $w$ and $r_0$ keep the profiles below $4\times10^{-6}A$ and $1.6\times10^{-4}A$, respectively, on the lines where the periodic distance is not smooth.}
\label{tab:exp_ic_families}
\begin{tabular}{llll}
\hline
Family & Initial distribution & Sampled parameters & Experiment case \\
\hline
\makecell[l]{I: Maxwellian \\ mixture}
& Eq.~\eqref{eq:2d_mixture_initial_general}
& \makecell[l]{$a_\rho\in[0.15,0.3]$, $a_g\in[0.1,0.2]$,\\ $w_x,w_y\in[0.3,0.6]$, $k\in\{1,2\}$,\\ $\varphi_x,\varphi_y\in[0,2\pi)$}
& Eq.~\eqref{eq:2d_mixture_initial}, $\sigma=2$ \\
\hline
\makecell[l]{II: Gaussian \\ pulse}
& Eq.~\eqref{eq:2d_blob_initial}
& \makecell[l]{$A\in[0.3,0.7]$, $w\in[0.06,0.1]$,\\ $\mathbf x_c\in[0.3,0.7]^2$}
& Eq.~\eqref{eq:2d_blob_test}, $\sigma=10$ \\
\hline
\makecell[l]{III: Ring \\ pulse}
& Eq.~\eqref{eq:2d_ring_initial}
& \makecell[l]{$A\in[0.3,0.7]$, $r_0\in[0.18,0.27]$,\\ $w\in[0.04,0.055]$, $\mathbf x_c=(0.5,0.5)$}
& Eq.~\eqref{eq:2d_ring_test}, $\sigma=8$ \\
\hline
Trajectories
& \multicolumn{3}{l}{$32$ (Family~I), $16$ (Families~II and~III), and the three experiment trajectories ($67$ in total)} \\
\hline
Collision frequency
& \multicolumn{3}{l}{$\log_{10}\sigma\sim\mathcal U(\log_{10}0.5,\log_{10}20)$, independently for each trajectory} \\
\hline
\end{tabular}
\end{table}

Regarding the network, Table~\ref{tab:hn_network} lists the architecture and the optimization settings. The shared symmetrizer network outputs the $d_3=31$ coefficients of the basis of $\mathcal L_3$ together with $\widehat\tau_3$, and each directional network outputs the $10$ independent entries of the symmetric $4\times4$ matrix $\widehat K_i$. Training minimizes Eq.~\eqref{eq:closure_training_loss} with $\gamma=0.02$ and $\epsilon_0=10^{-12}$ by Algorithm~\ref{alg:hn_training}. The inverse symmetrizer uses $\eta=0.9$ and $\varepsilon=10^{-4}$, and the bounded completion uses $c_0=3.8897$ and $\delta_c=0.02$. The final checkpoint is used for moment evolution in all three experiments.

\begin{table}[htbp!]
\footnotesize
\renewcommand{\arraystretch}{1.08}
\setlength{\tabcolsep}{4pt}
\centering
\caption{Neural network architecture and optimization settings of the HN closure, shared by all three numerical experiments (Secs.~\ref{ex1_2d}, \ref{ex2_2d}, and~\ref{ex3_2d}). }
\label{tab:hn_network}
\begin{tabular}{lccc}
\hline
 & \makecell[c]{Symmetrizer \\ network}
 & \makecell[c]{Directional \\ network ($x$)}
 & \makecell[c]{Directional \\ network ($y$)} \\
\hline
Input dimension & $10$ & $10$ & $10$ \\
Output dimension & $d_3+1=32$ & $10$ & $10$ \\
Hidden layers & $[512]\times5$ & $[512]\times5$ & $[512]\times5$ \\
Activation & SiLU & SiLU & SiLU \\
Trainable parameters & $1{,}072{,}672$ & $1{,}061{,}386$ & $1{,}061{,}386$ \\
\hline
Total trainable parameters & \multicolumn{3}{c}{$3{,}195{,}444$} \\
\hline
Optimizer & \multicolumn{3}{c}{Adam, batch size $4096$} \\
Training samples & \multicolumn{3}{c}{$1.61\times10^7$ ($80\%$ of $67$ trajectories $\times\,3\times10^5$ samples)} \\
Epochs & \multicolumn{3}{c}{$320$ ($1.26\times10^6$ iterations)} \\
Learning rate & \multicolumn{3}{c}{\makecell[c]{epochs $1$--$160$: $10^{-3}$, multiplied by $0.35$ every $40$ epochs;\\ epochs $161$--$320$: $3\times10^{-4}$, multiplied by $0.4$ every $40$ epochs}} \\
Structural parameters & \multicolumn{3}{c}{$\eta=0.9$, $\varepsilon=10^{-4}$, $c_0=3.8897$, $\delta_c=0.02$} \\
Loss scales & \multicolumn{3}{c}{$\gamma=0.02$, $\epsilon_0=10^{-12}$} \\
\hline
\end{tabular}
\end{table}

\begin{figure}[htp!]
\centering
\includegraphics[width=0.6\linewidth]{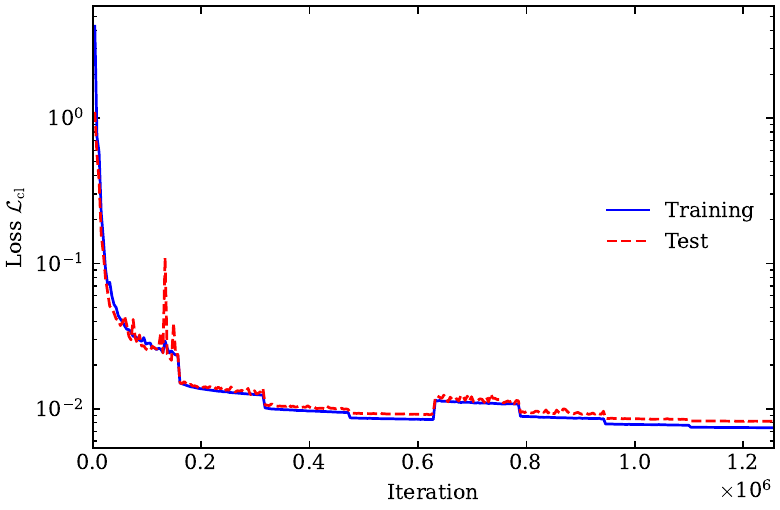}
\caption{Training history of the HN closure: the loss in Eq.~\eqref{eq:closure_training_loss} on the training and test sets. One epoch corresponds to $3{,}926$ iterations with batch size $4096$.}
\label{fig:closure_loss_history}
\end{figure}

Figure~\ref{fig:closure_loss_history} shows the training history. The loss decreases by more than two orders of magnitude, mostly in the first $1.6\times10^5$ iterations ($40$ epochs), and follows the learning-rate schedule in Table~\ref{tab:hn_network}. The training and test losses remain close throughout, with no sign of overfitting.

\begin{figure}[htp!]
\centering
\includegraphics[width=\linewidth]{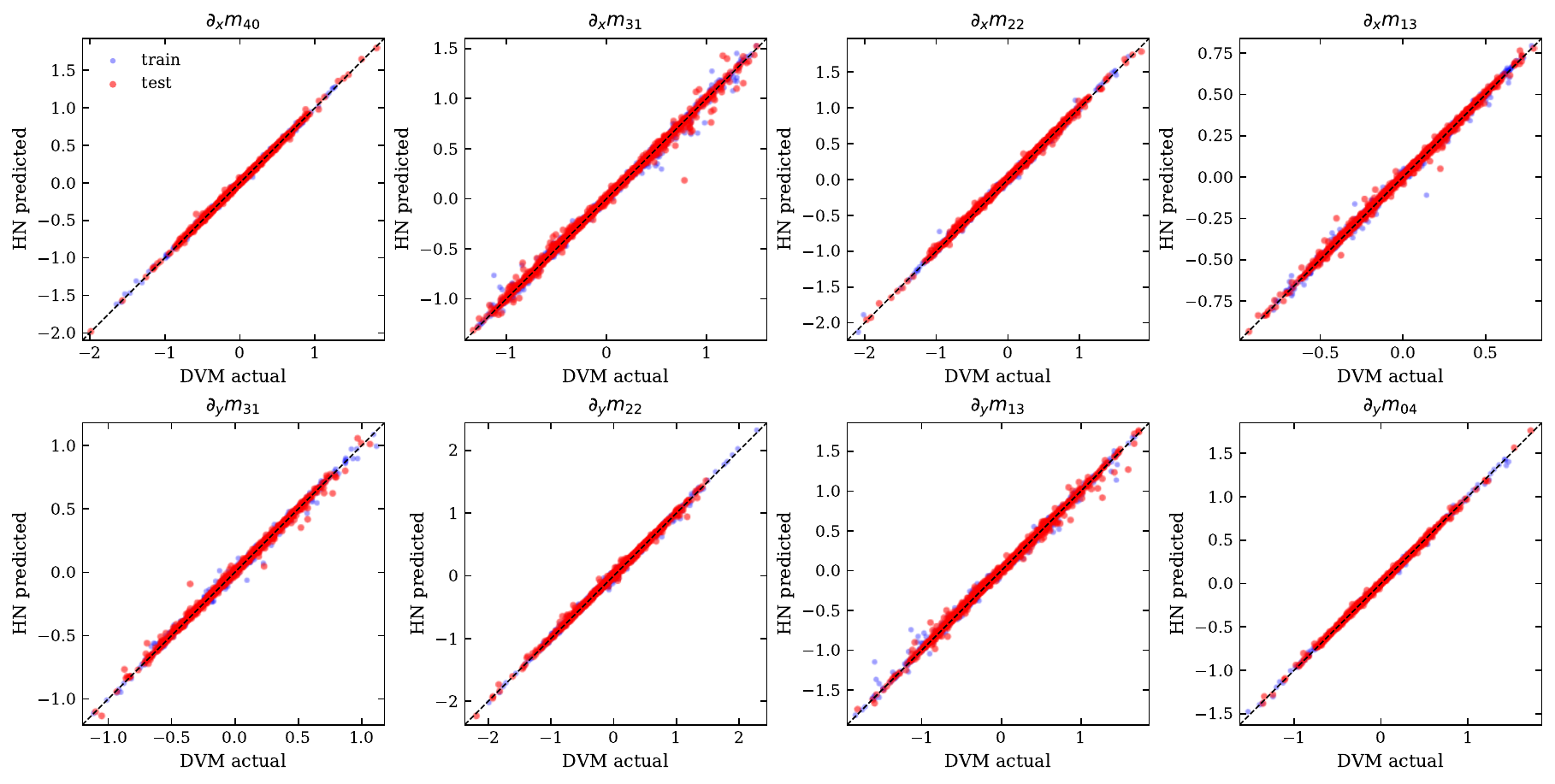}
\caption{Unresolved derivatives predicted by the HN closure versus the DVM labels, for randomly selected training (blue) and test (red) samples. The top row shows the $x$-derivatives $\partial_xm_{40}$, $\partial_xm_{31}$, $\partial_xm_{22}$, and $\partial_xm_{13}$, and the bottom row the $y$-derivatives $\partial_ym_{31}$, $\partial_ym_{22}$, $\partial_ym_{13}$, and $\partial_ym_{04}$. The dashed line indicates exact agreement.}
\label{fig:closure_actual_vs_predicted}
\end{figure}

Figure~\ref{fig:closure_actual_vs_predicted} compares the predicted unresolved derivatives in Eq.~\eqref{eq:unresolved_gradient_predictions} with the DVM labels in Eq.~\eqref{eq:unresolved_gradient_labels} for randomly selected training and test samples. All eight components lie close to the diagonal, and the training and test samples are indistinguishable; after training, the loss in Eq.~\eqref{eq:closure_training_loss} is $7.4\times10^{-3}$ on the training set and $8.2\times10^{-3}$ on the test set. On the three experiment trajectories, the relative $L^2$ error of the predicted derivatives, $\|\widehat{\mathbf g}_i-\mathbf g_i\|_{2,h}/\|\mathbf g_i\|_{2,h}$, is at most $2.7\%$, $1.3\%$, and $1.8\%$ in Experiments~\ref{ex1_2d}, \ref{ex2_2d}, and~\ref{ex3_2d}, respectively, for $0.1\le t\le0.5$ and $i\in\{x,y\}$. Thus, a single closure represents the unresolved fluxes of all three cases with comparable accuracy, even though the matrices are constrained to the hyperbolic form of Section~\ref{sec:neural_parameterization}.

\begin{figure}[htp!]
\centering
\includegraphics[width=\linewidth]{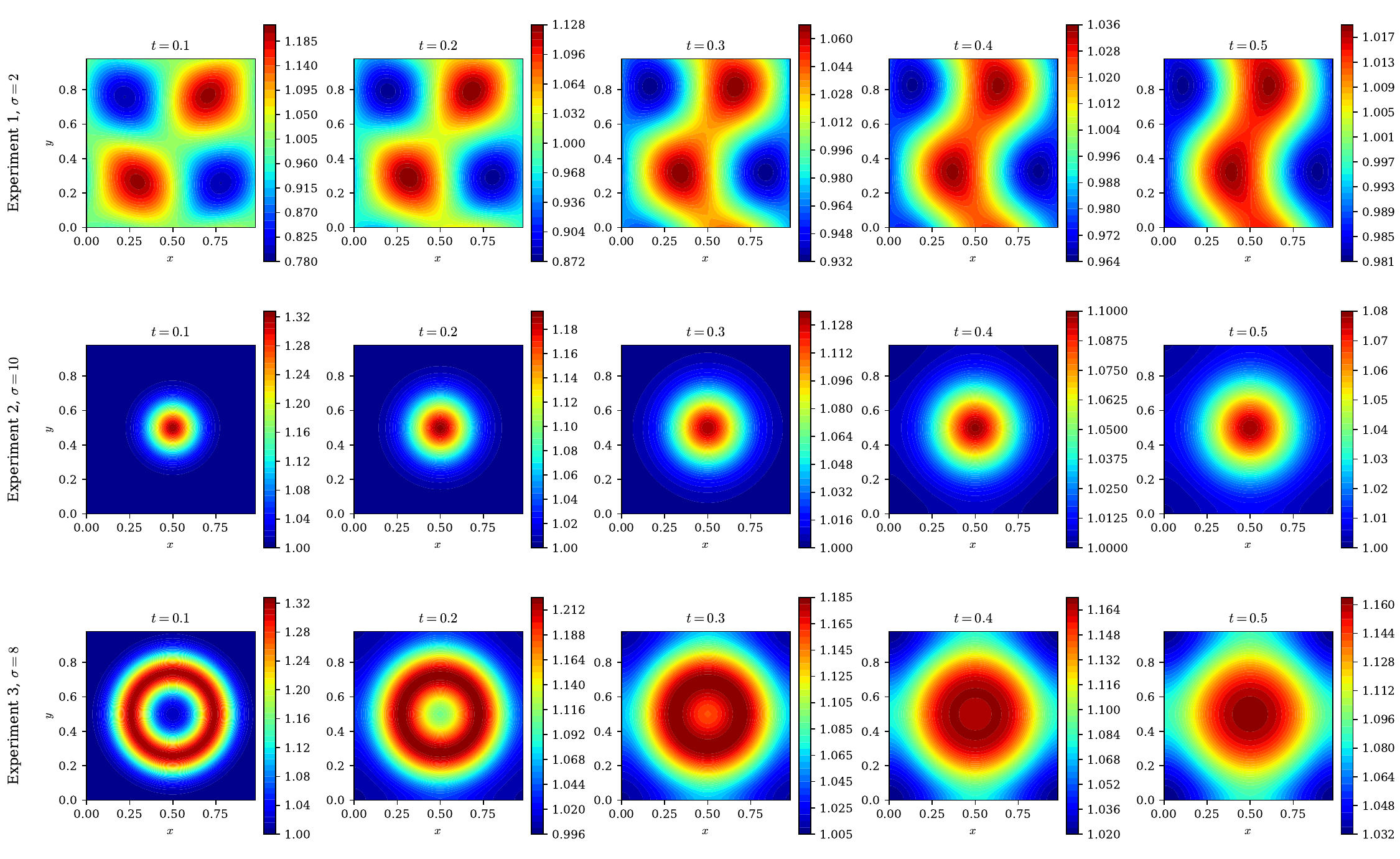}
\caption{DVM reference densities of the three experiment cases. Rows correspond to Experiment~\ref{ex1_2d} (Maxwellian mixture, $\sigma=2$), Experiment~\ref{ex2_2d} (Gaussian pulse, $\sigma=10$), and Experiment~\ref{ex3_2d} (ring pulse, $\sigma=8$), and columns to $t=0.1$, $0.2$, $0.3$, $0.4$, and $0.5$.}
\label{fig:dvm_overview}
\end{figure}

To illustrate the dynamics that the closure must represent, Fig.~\ref{fig:dvm_overview} shows the DVM densities of the three experiment cases. The three rows display qualitatively different behavior: relaxation of a non-equilibrium mixture, diffusive spreading of a pulse, and inward convergence of a ring into a single central peak.

\subsection{Experiment 1: Maxwellian mixture in the transition regime}
\label{ex1_2d}

We first test the proposed closure on a smooth periodic problem whose initial distribution is not a local equilibrium. This example serves as a benchmark for both closure accuracy and the stability of the partially conservative DG solver. The kinetic equation is Eq.~\eqref{eq:2d_bte} with
\begin{equation}
\frac{\partial f}{\partial t}+v_x\frac{\partial f}{\partial x}+v_y\frac{\partial f}{\partial y}
=\sigma\bigl[M_2(\mathbf v)\rho_f-f\bigr],
\qquad
\sigma=2,
\qquad
(\mathbf x,t)\in[0,1]^2\times(0,T],
\label{eq:ex1_kinetic}
\end{equation}
with periodic boundary conditions in $x$ and $y$. The corresponding moment system is Eq.~\eqref{eq:exp_closed_system} with $\sigma=2$, i.e., $\mathrm{Kn}=0.5$.

The training family (Family~I in Table~\ref{tab:exp_ic_families}) is a positive mixture of a centered Maxwellian and a Maxwellian shifted in velocity by a spatially varying vector,
\begin{align}
 f_0(\mathbf x,\mathbf v)
 &=\rho_0(\mathbf x)\left[(1-g(\mathbf x))M_2(\mathbf v)
                 +g(\mathbf x)M_2(\mathbf v-\mathbf w(\mathbf x))\right],\nonumber\\
 \rho_0(x,y)&=1+a_\rho\sin(2\pi kx+\varphi_x)\sin(2\pi ky+\varphi_y),\nonumber\\
 g(x,y)&=0.3+a_g\sin(2\pi kx+\varphi_x)\sin(2\pi ky+\varphi_y),\nonumber\\
 \mathbf w(x,y)&=\bigl(w_x\sin(2\pi ky+\varphi_y),\ w_y\cos(2\pi kx+\varphi_x)\bigr)^{\mathsf T}.
 \label{eq:2d_mixture_initial_general}
\end{align}
The case of Experiment~\ref{ex1_2d} is
\begin{align}
 \rho_0(x,y)&=1+0.25\sin(2\pi x)\sin(2\pi y),\qquad
 g(x,y)=0.3+0.15\sin(2\pi x)\sin(2\pi y),\nonumber\\
 \mathbf w(x,y)&=(0.5\sin(2\pi y),\ 0.4\cos(2\pi x))^{\mathsf T},
 \label{eq:2d_mixture_initial}
\end{align}
that is, $a_\rho=0.25$, $a_g=0.15$, $w_x=0.5$, $w_y=0.4$, $k=1$, and $\varphi_x=\varphi_y=0$. Since the Hermite moments of the shifted Maxwellian $M_2(\mathbf v-\mathbf w)$ are generically nonzero in every degree, the initial distribution is not a local equilibrium. The varying density and velocity shifts therefore activate several spatial and Hermite modes from $t=0$. Thus, this example tests whether the closure follows the relaxation of an initially non-equilibrium state while transport and collisions are of comparable strength.

\begin{figure}[htp!]
\centering
\includegraphics[width=\linewidth,height=0.9\textheight,keepaspectratio]{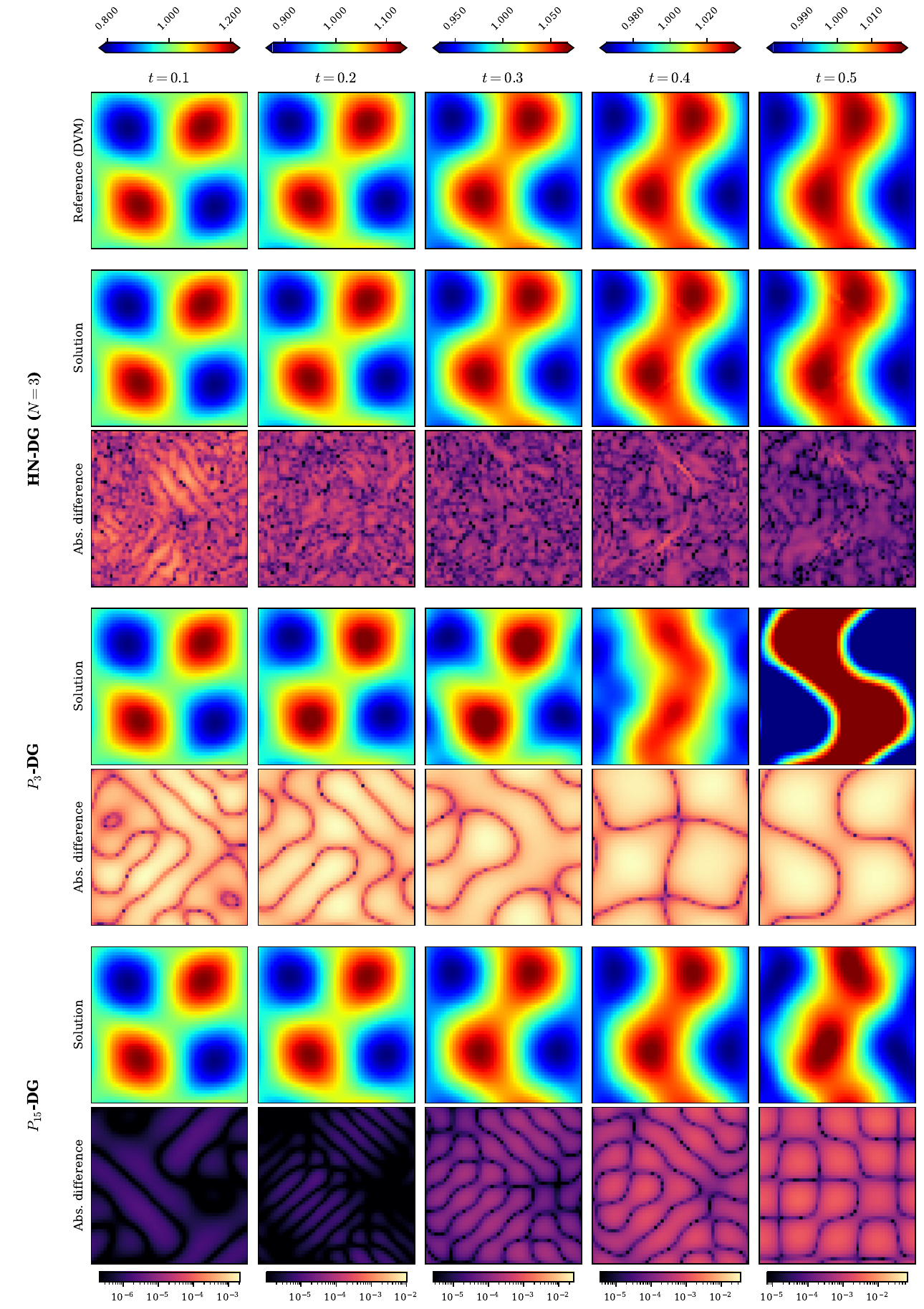}
\caption{Density of the Maxwellian mixture in Experiment~\ref{ex1_2d} ($\sigma=2$) at $t=0.1$, $0.2$, $0.3$, $0.4$, and $0.5$. The first row shows the DVM reference; each method then occupies two rows, its solution and the absolute difference from the reference, for the proposed HN-DG with $N=3$ and the classical $P_3$ and $P_{15}$ truncations; the errors of $P_7$ are listed in the corresponding table. In each column, the solutions use the color range of the reference, and the absolute differences use a common logarithmic scale.}
\label{fig:ex1_density}
\end{figure}

Figure~\ref{fig:ex1_density} compares the densities of the DVM reference, the HN closure, and the classical truncations, and Table~\ref{tab:ex1_errors} lists the corresponding errors. For $t\le0.2$, all methods except $P_3$ remain close to the reference; $P_7$ is more accurate than HN-DG at $t=0.1$, and $P_{15}$ up to $t=0.3$. As the mixture relaxes, the errors of the classical truncations grow and form coherent patterns that follow the level sets of the density. The density variation of $P_3$ decays much more slowly than in the reference, and at $t=0.5$ its error is three times the variation itself. The error of HN-DG grows more slowly and shows no coherent structure. At $t=0.5$, $E_{\delta\rho}$ is $2.1\%$ for HN-DG, compared with $50\%$ for $P_7$ and $11.9\%$ for $P_{15}$. Thus, with the same ten moments as $P_3$, the HN closure is more accurate than $P_{15}$, which uses $136$ moments, for $t\ge0.4$.

With the same DG implementation, HN-DG requires about $290$\,s to reach $T=0.5$, compared with $9$\,s for $P_3$, $61$\,s for $P_7$, and $622$\,s for $P_{15}$. The additional cost relative to $P_3$ comes from the network evaluations and the spectral radii $\rho(B_n)$ at every quadrature point and stage; this part of the implementation was not optimized. The same timing pattern holds in the other two experiments.

\begin{table}[htp!]
\footnotesize
\centering
\setlength{\tabcolsep}{3.5pt}
\caption{Density errors in Eq.~\eqref{eq:density_error_metrics} for the Maxwellian mixture in Experiment~\ref{ex1_2d}, at the times shown in Fig.~\ref{fig:ex1_density}. The last column lists the wall-clock time of the DG solver to reach $T=0.5$, excluding output.}
\label{tab:ex1_errors}
\begin{tabular}{lcccccccc}
\hline
Method & Moments & Error & $t=0.1$ & $t=0.2$ & $t=0.3$ & $t=0.4$ & $t=0.5$ & Time (s) \\
\hline
\multirow{2}{*}{HN-DG ($N=3$)} & \multirow{2}{*}{$10$} & $E_\rho$ & $7.23\times10^{-5}$ & $1.08\times10^{-4}$ & $1.50\times10^{-4}$ & $2.37\times10^{-4}$ & $2.38\times10^{-4}$ & \multirow{2}{*}{$288$} \\
 & & $E_{\delta\rho}$ & $7.03\times10^{-4}$ & $1.68\times10^{-3}$ & $4.05\times10^{-3}$ & $1.14\times10^{-2}$ & $2.06\times10^{-2}$ & \\
\hline
\multirow{2}{*}{$P_3$} & \multirow{2}{*}{$10$} & $E_\rho$ & $7.34\times10^{-4}$ & $4.18\times10^{-3}$ & $9.96\times10^{-3}$ & $1.58\times10^{-2}$ & $3.55\times10^{-2}$ & \multirow{2}{*}{$9$} \\
 & & $E_{\delta\rho}$ & $7.14\times10^{-3}$ & $6.47\times10^{-2}$ & $2.69\times10^{-1}$ & $7.65\times10^{-1}$ & $3.06\times10^{0}$ & \\
\hline
\multirow{2}{*}{$P_7$} & \multirow{2}{*}{$36$} & $E_\rho$ & $3.71\times10^{-6}$ & $2.70\times10^{-4}$ & $1.82\times10^{-3}$ & $3.61\times10^{-3}$ & $5.84\times10^{-3}$ & \multirow{2}{*}{$61$} \\
 & & $E_{\delta\rho}$ & $3.60\times10^{-5}$ & $4.18\times10^{-3}$ & $4.91\times10^{-2}$ & $1.75\times10^{-1}$ & $5.03\times10^{-1}$ & \\
\hline
\multirow{2}{*}{$P_{15}$} & \multirow{2}{*}{$136$} & $E_\rho$ & $9.20\times10^{-7}$ & $3.00\times10^{-6}$ & $7.87\times10^{-5}$ & $3.97\times10^{-4}$ & $1.38\times10^{-3}$ & \multirow{2}{*}{$622$} \\
 & & $E_{\delta\rho}$ & $8.95\times10^{-6}$ & $4.63\times10^{-5}$ & $2.13\times10^{-3}$ & $1.92\times10^{-2}$ & $1.19\times10^{-1}$ & \\
\hline
\end{tabular}
\end{table}

\subsection{Experiment 2: Gaussian pulse in the scattering-dominated regime}
\label{ex2_2d}

We next test the proposed closure on a localized density pulse in a strongly scattering medium. Unlike Experiment~\ref{ex1_2d}, the initial distribution is a local equilibrium with no bulk velocity, so all non-equilibrium moments are generated by transport. The kinetic equation is
\begin{equation}
\frac{\partial f}{\partial t}+v_x\frac{\partial f}{\partial x}+v_y\frac{\partial f}{\partial y}
=\sigma\bigl[M_2(\mathbf v)\rho_f-f\bigr],
\qquad
\sigma=10,
\qquad
(\mathbf x,t)\in[0,1]^2\times(0,T],
\label{eq:ex2_kinetic}
\end{equation}
with periodic boundary conditions, i.e., $\mathrm{Kn}=0.1$. The moment system is Eq.~\eqref{eq:exp_closed_system} with $\sigma=10$.

For a center $\mathbf x_c\in\Omega$, let
\[
d(\mathbf x,\mathbf x_c):=\min_{\mathbf k\in\mathbb Z^2}|\mathbf x-\mathbf x_c-\mathbf k|
\]
denote the periodic distance. The training family (Family~II in Table~\ref{tab:exp_ic_families}) is an isotropic Gaussian density pulse on a uniform background,
\begin{equation}
f_0(\mathbf x,\mathbf v)=\rho_0(\mathbf x)M_2(\mathbf v),
\qquad
\rho_0(\mathbf x)=1+A\exp\left(-\frac{d(\mathbf x,\mathbf x_c)^2}{2w^2}\right).
\label{eq:2d_blob_initial}
\end{equation}
The case of Experiment~\ref{ex2_2d} is
\begin{equation}
A=0.5,\qquad w=0.08,\qquad \mathbf x_c=(0.5,0.5).
\label{eq:2d_blob_test}
\end{equation}
The periodic distance is not smooth on the lines at distance $1/2$ from $\mathbf x_c$, where the test pulse is smaller than $10^{-8}$, so the periodic distance introduces no visible kink. Since $f_0=\rho_0M_2$, the initial state satisfies
\[
m_{00}(\mathbf x,0)=\rho_0(\mathbf x),
\qquad
m_{ab}(\mathbf x,0)=0,\quad 1\le a+b\le 3,
\]
and the initial unresolved moments vanish as well.

For $\sigma\gg1$, a Chapman--Enskog expansion of Eq.~\eqref{eq:ex2_kinetic} gives, to leading order, the diffusion equation
\begin{equation}
\frac{\partial\rho}{\partial t}=D\,\Delta\rho,
\qquad
D=\frac{1}{\sigma}\int_{\R^2}v_x^2M_2(\mathbf v)\,d\mathbf v=\frac{1}{2\sigma},
\label{eq:ex2_diffusion_limit}
\end{equation}
whose solution for the isotropic pulse, before periodic images interact, is
\begin{equation}
\rho_{\mathrm{diff}}(\mathbf x,t)
=1+\frac{Aw^2}{w^2+2Dt}
\exp\left(-\frac{d(\mathbf x,\mathbf x_c)^2}{2(w^2+2Dt)}\right).
\label{eq:ex2_diffusion_solution}
\end{equation}
With $\sigma=10$, $D=0.05$, and the pulse width grows from $0.08$ to approximately $0.24$ at $T=0.5$. Equation~\eqref{eq:ex2_diffusion_solution} is not the reference solution, since $\mathrm{Kn}=0.1$ retains kinetic corrections, but it indicates the asymptotic behavior that the closed moment system must recover. Thus, this example tests whether the closure preserves the scattering-dominated limit, where the moment hierarchy should reduce to diffusive transport without spurious wave fronts.

\begin{figure}[htp!]
\centering
\includegraphics[width=\linewidth,height=0.9\textheight,keepaspectratio]{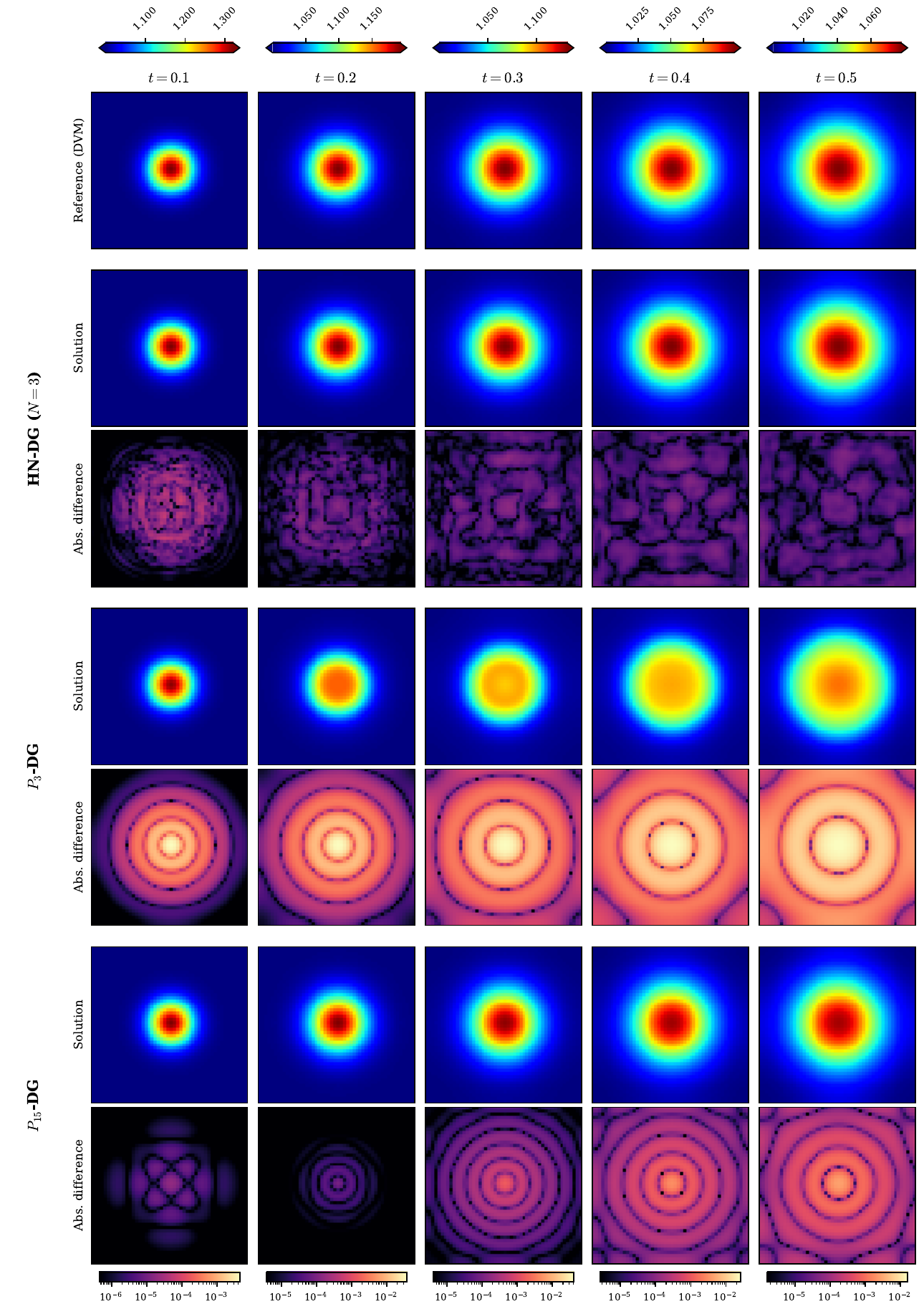}
\caption{Density of the Gaussian pulse in Experiment~\ref{ex2_2d} ($\sigma=10$) at $t=0.1$, $0.2$, $0.3$, $0.4$, and $0.5$. The first row shows the DVM reference; each method then occupies two rows, its solution and the absolute difference from the reference, for the proposed HN-DG with $N=3$ and the classical $P_3$ and $P_{15}$ truncations; the errors of $P_7$ are listed in the corresponding table. In each column, the solutions use the color range of the reference, and the absolute differences use a common logarithmic scale.}
\label{fig:ex2_density}
\end{figure}

Figure~\ref{fig:ex2_density} compares the densities of the DVM reference, the HN closure, and the classical truncations, and Table~\ref{tab:ex2_errors} lists the corresponding errors. HN-DG keeps $E_{\delta\rho}$ below $0.14\%$ at all times, and its error has no radial structure. The errors of the classical truncations form concentric rings, which correspond to wave fronts that travel at the finite characteristic speeds of the truncated systems. $P_3$ additionally flattens the peak of the pulse for $0.2\le t\le0.4$. At $t=0.5$, $E_{\delta\rho}$ is $21\%$, $7.3\%$, and $2.0\%$ for $P_3$, $P_7$, and $P_{15}$, respectively, and $0.08\%$ for HN-DG. $P_{15}$ is more accurate than HN-DG only for $t\le0.2$. Therefore, the learned third-order system recovers the diffusive behavior of the scattering-dominated regime without spurious fronts, whereas the classical truncations approach it only slowly as the order increases.

\begin{table}[htp!]
\footnotesize
\centering
\setlength{\tabcolsep}{3.5pt}
\caption{Density errors in Eq.~\eqref{eq:density_error_metrics} for the Gaussian pulse in Experiment~\ref{ex2_2d}, at the times shown in Fig.~\ref{fig:ex2_density}. The last column lists the wall-clock time of the DG solver to reach $T=0.5$, excluding output.}
\label{tab:ex2_errors}
\begin{tabular}{lcccccccc}
\hline
Method & Moments & Error & $t=0.1$ & $t=0.2$ & $t=0.3$ & $t=0.4$ & $t=0.5$ & Time (s) \\
\hline
\multirow{2}{*}{HN-DG ($N=3$)} & \multirow{2}{*}{$10$} & $E_\rho$ & $1.02\times10^{-5}$ & $3.19\times10^{-5}$ & $3.81\times10^{-5}$ & $2.78\times10^{-5}$ & $1.50\times10^{-5}$ & \multirow{2}{*}{$294$} \\
 & & $E_{\delta\rho}$ & $1.97\times10^{-4}$ & $8.49\times10^{-4}$ & $1.32\times10^{-3}$ & $1.22\times10^{-3}$ & $8.14\times10^{-4}$ & \\
\hline
\multirow{2}{*}{$P_3$} & \multirow{2}{*}{$10$} & $E_\rho$ & $3.71\times10^{-4}$ & $4.03\times10^{-3}$ & $5.73\times10^{-3}$ & $5.01\times10^{-3}$ & $3.89\times10^{-3}$ & \multirow{2}{*}{$9$} \\
 & & $E_{\delta\rho}$ & $7.16\times10^{-3}$ & $1.07\times10^{-1}$ & $1.99\times10^{-1}$ & $2.20\times10^{-1}$ & $2.11\times10^{-1}$ & \\
\hline
\multirow{2}{*}{$P_7$} & \multirow{2}{*}{$36$} & $E_\rho$ & $2.84\times10^{-6}$ & $4.88\times10^{-4}$ & $1.55\times10^{-3}$ & $1.73\times10^{-3}$ & $1.35\times10^{-3}$ & \multirow{2}{*}{$61$} \\
 & & $E_{\delta\rho}$ & $5.47\times10^{-5}$ & $1.30\times10^{-2}$ & $5.38\times10^{-2}$ & $7.58\times10^{-2}$ & $7.31\times10^{-2}$ & \\
\hline
\multirow{2}{*}{$P_{15}$} & \multirow{2}{*}{$136$} & $E_\rho$ & $2.19\times10^{-6}$ & $9.16\times10^{-6}$ & $1.74\times10^{-4}$ & $3.75\times10^{-4}$ & $3.71\times10^{-4}$ & \multirow{2}{*}{$622$} \\
 & & $E_{\delta\rho}$ & $4.23\times10^{-5}$ & $2.44\times10^{-4}$ & $6.05\times10^{-3}$ & $1.64\times10^{-2}$ & $2.01\times10^{-2}$ & \\
\hline
\end{tabular}
\end{table}

\subsection{Experiment 3: Converging ring pulse in the scattering-dominated regime}
\label{ex3_2d}

We finally consider a ring-shaped pulse whose inward-moving part converges toward the center of the ring in a strongly scattering medium. In contrast to the spreading pulse of Experiment~\ref{ex2_2d}, the density at the center first increases, so that the solution passes through a configuration in which inward and outward fluxes overlap. The kinetic equation is
\begin{equation}
\frac{\partial f}{\partial t}+v_x\frac{\partial f}{\partial x}+v_y\frac{\partial f}{\partial y}
=\sigma\bigl[M_2(\mathbf v)\rho_f-f\bigr],
\qquad
\sigma=8,
\qquad
(\mathbf x,t)\in[0,1]^2\times(0,T],
\label{eq:ex3_kinetic}
\end{equation}
with periodic boundary conditions, i.e., $\mathrm{Kn}=0.125$. The moment system is Eq.~\eqref{eq:exp_closed_system} with $\sigma=8$.

The training family (Family~III in Table~\ref{tab:exp_ic_families}) is an annular density pulse centered at $\mathbf x_c=(0.5,0.5)$,
\begin{equation}
f_0(\mathbf x,\mathbf v)=\rho_0(\mathbf x)M_2(\mathbf v),
\qquad
\rho_0(\mathbf x)=1+A\exp\left(-\frac{\bigl(d(\mathbf x,\mathbf x_c)-r_0\bigr)^2}{2w^2}\right),
\label{eq:2d_ring_initial}
\end{equation}
and the case of Experiment~\ref{ex3_2d} is
\begin{equation}
A=0.5,\qquad r_0=0.25,\qquad w=0.05.
\label{eq:2d_ring_test}
\end{equation}
As in Experiment~\ref{ex2_2d}, the initial distribution is a local equilibrium with vanishing non-equilibrium moments. Particles leaving the ring with inward velocities converge toward $\mathbf x_c$, while those with outward velocities spread toward the periodic boundary. In the collisionless limit, the focusing time is of order
\[
t_{\mathrm f}\approx\frac{r_0}{\langle|\mathbf v|\rangle},
\qquad
\langle|\mathbf v|\rangle=\int_{\R^2}|\mathbf v|M_2(\mathbf v)\,d\mathbf v=\frac{\sqrt\pi}{2},
\]
which gives $t_{\mathrm f}\approx0.28$ for $r_0=0.25$. With $\sigma=8$, the mean free time $\sigma^{-1}=0.125$ is shorter than $t_{\mathrm f}$, so scattering randomizes the inward velocities before they reach the center and delays the convergence. In the DVM reference, the density at the center is still lower than on the ring at $t=0.2$ and $0.3$, the profile becomes flat at $t\approx0.4$, and the focal density reaches its maximum $\rho_c=1.1656$ at $t=0.41$; it decreases only slightly to $\rho_c=1.1624$ at $T=0.5$. The density variation $\|\rho_{\mathrm{DVM}}-\langle\rho_{\mathrm{DVM}}\rangle\|_{2,h}$ at $T=0.5$ is $23\%$ of its initial value. The whole convergence process therefore takes place within the simulated interval. While the inward front passes through the center, particles arrive from all directions and the velocity distribution near $\mathbf x_c$ departs from a Maxwellian, whereas scattering keeps this departure moderate. Thus, this example tests the closure on a radially converging flow, which is absent from Experiments~\ref{ex1_2d} and~\ref{ex2_2d}.

\begin{figure}[htp!]
\centering
\includegraphics[width=\linewidth,height=0.9\textheight,keepaspectratio]{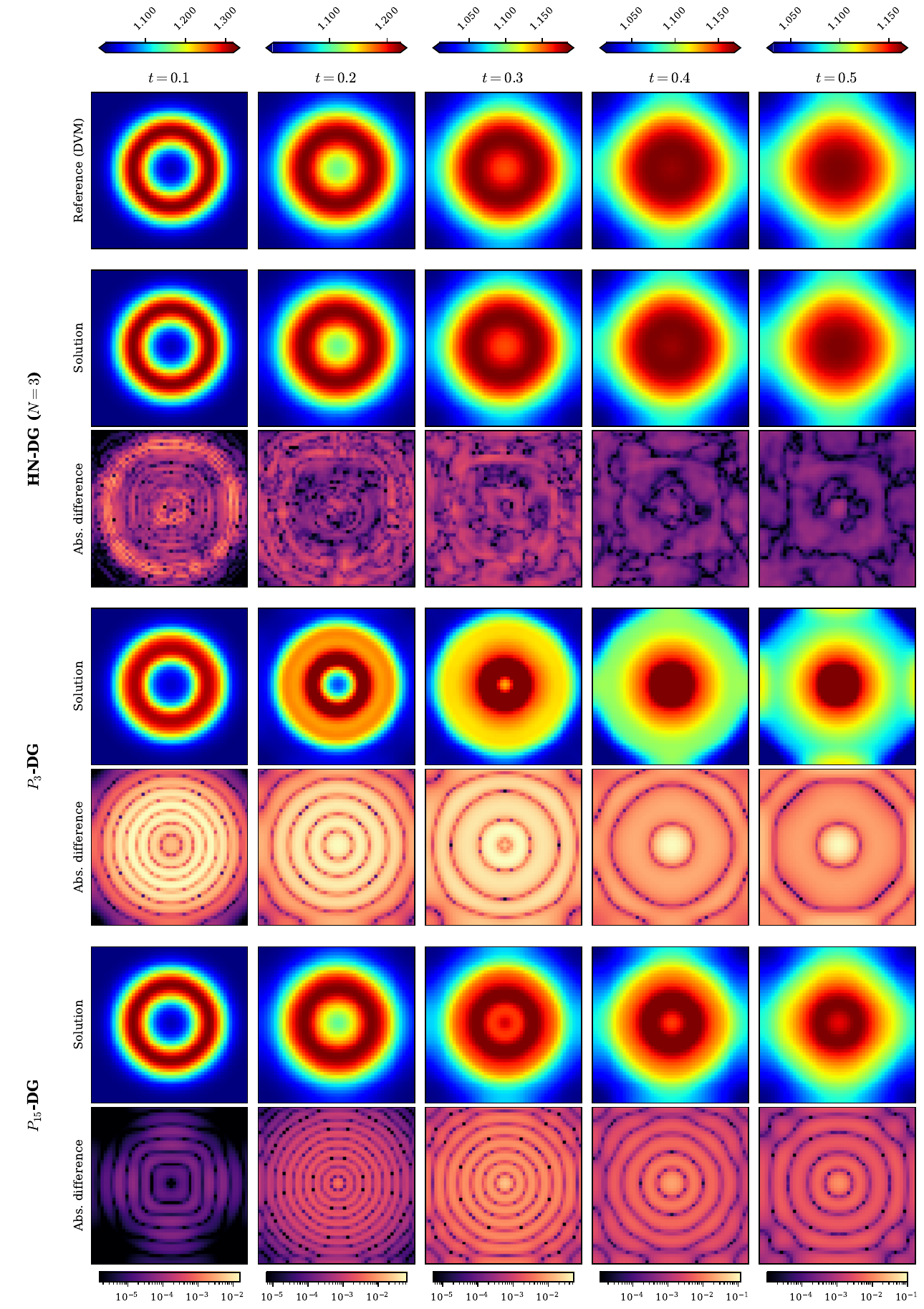}
\caption{Density of the converging ring pulse in Experiment~\ref{ex3_2d} ($\sigma=8$) at $t=0.1$, $0.2$, $0.3$, $0.4$, and $0.5$. The first row shows the DVM reference; each method then occupies two rows, its solution and the absolute difference from the reference, for the proposed HN-DG with $N=3$ and the classical $P_3$ and $P_{15}$ truncations; the errors of $P_7$ are listed in the corresponding table. In each column, the solutions use the color range of the reference, and the absolute differences use a common logarithmic scale.}
\label{fig:ex3_density}
\end{figure}

Figure~\ref{fig:ex3_density} compares the densities of the DVM reference, the HN closure, and the classical truncations, and Table~\ref{tab:ex3_errors} lists the corresponding errors. At $t=0.1$, before the front reaches the center, $P_7$ and $P_{15}$ are more accurate than HN-DG. As the ring converges, the classical truncations produce artificial structures at the focal point: $P_3$ shows a central spot at $t=0.2$ and $0.3$ that is absent from the reference, and the error of $P_{15}$ forms concentric rings around the center. HN-DG reproduces the filling of the center and the formation of the focal maximum. Its $E_{\delta\rho}$ remains below $0.84\%$ at all times and is $0.62\%$ at $t=0.5$, compared with $43\%$, $16\%$, and $7.2\%$ for $P_3$, $P_7$, and $P_{15}$. Thus, the closure remains accurate for a radially converging flow, in which the velocity distribution near the center is not aligned with either coordinate direction.

\begin{table}[htp!]
\footnotesize
\centering
\setlength{\tabcolsep}{3.5pt}
\caption{Density errors in Eq.~\eqref{eq:density_error_metrics} for the converging ring pulse in Experiment~\ref{ex3_2d}, at the times shown in Fig.~\ref{fig:ex3_density}. The last column lists the wall-clock time of the DG solver to reach $T=0.5$, excluding output.}
\label{tab:ex3_errors}
\begin{tabular}{lcccccccc}
\hline
Method & Moments & Error & $t=0.1$ & $t=0.2$ & $t=0.3$ & $t=0.4$ & $t=0.5$ & Time (s) \\
\hline
\multirow{2}{*}{HN-DG ($N=3$)} & \multirow{2}{*}{$10$} & $E_\rho$ & $3.46\times10^{-4}$ & $5.64\times10^{-4}$ & $4.35\times10^{-4}$ & $3.03\times10^{-4}$ & $2.06\times10^{-4}$ & \multirow{2}{*}{$308$} \\
 & & $E_{\delta\rho}$ & $3.43\times10^{-3}$ & $8.15\times10^{-3}$ & $8.33\times10^{-3}$ & $7.33\times10^{-3}$ & $6.19\times10^{-3}$ & \\
\hline
\multirow{2}{*}{$P_3$} & \multirow{2}{*}{$10$} & $E_\rho$ & $5.18\times10^{-3}$ & $2.07\times10^{-2}$ & $1.88\times10^{-2}$ & $1.74\times10^{-2}$ & $1.43\times10^{-2}$ & \multirow{2}{*}{$9$} \\
 & & $E_{\delta\rho}$ & $5.14\times10^{-2}$ & $2.99\times10^{-1}$ & $3.60\times10^{-1}$ & $4.21\times10^{-1}$ & $4.29\times10^{-1}$ & \\
\hline
\multirow{2}{*}{$P_7$} & \multirow{2}{*}{$36$} & $E_\rho$ & $2.88\times10^{-4}$ & $6.68\times10^{-3}$ & $1.02\times10^{-2}$ & $6.86\times10^{-3}$ & $5.36\times10^{-3}$ & \multirow{2}{*}{$61$} \\
 & & $E_{\delta\rho}$ & $2.85\times10^{-3}$ & $9.65\times10^{-2}$ & $1.95\times10^{-1}$ & $1.66\times10^{-1}$ & $1.61\times10^{-1}$ & \\
\hline
\multirow{2}{*}{$P_{15}$} & \multirow{2}{*}{$136$} & $E_\rho$ & $1.69\times10^{-5}$ & $9.62\times10^{-4}$ & $3.17\times10^{-3}$ & $3.61\times10^{-3}$ & $2.39\times10^{-3}$ & \multirow{2}{*}{$622$} \\
 & & $E_{\delta\rho}$ & $1.67\times10^{-4}$ & $1.39\times10^{-2}$ & $6.06\times10^{-2}$ & $8.74\times10^{-2}$ & $7.19\times10^{-2}$ & \\
\hline
\end{tabular}
\end{table}

\section{Conclusion}
\label{sec:conclusion}

We developed a hyperbolic neural closure for multidimensional Hermite moment systems of the linear Boltzmann equation. To the best of our knowledge, it is the first data-driven closure that guarantees hyperbolicity of a multidimensional moment system of the Boltzmann equation for every network output; previous learned closures with this guarantee for this equation were formulated in one spatial dimension. The closure learns an inverse symmetrizer within the subspace determined by the known Hermite transport rows. A normalized perturbation of the identity keeps this symmetrizer positive definite, and symmetric completion yields the unresolved rows. Because the same symmetrizer is shared by both spatial directions, every directional Jacobian has real eigenvalues and a complete set of eigenvectors, and the known lower-order equations remain exact. These properties hold for every network output and at every truncation order.

The closure is trained only on spatial derivatives of unresolved moments from kinetic reference solutions, with no solver in the training loop. For moment evolution, the lower-order equations are discretized with conservative DG fluxes, and the highest-order equations with volume products and path fluctuations; SSPRK3 transport steps are combined with exact collision updates by Strang splitting. In three two-dimensional experiments covering the transition and scattering-dominated regimes, the learned third-order system remained stable, and a single closure gave time-averaged relative errors in the density variation of $0.77\%$, $0.09\%$, and $0.67\%$ over $t=0.1,0.2,\ldots,0.5$, compared with $83\%$, $15\%$, and $31\%$ for $P_3$. In all three experiments, the ten-moment system was also more accurate, averaged over time, than the $P_7$ and $P_{15}$ truncations with $36$ and $136$ moments. Hyperbolicity by construction therefore allows a low-order learned system to be used in a DG solver and to compete with much larger classical moment systems.

Several extensions remain. The present construction assumes a fixed equilibrium and a linear collision operator; extending it to the nonlinear BGK and Boltzmann operators would require a state-dependent Hermite basis. Boundary conditions, higher truncation orders, and realizability of the underlying distribution were not addressed and are left for future work.

\newpage

\end{document}